\documentclass[journal]{IEEEtran}
\IEEEoverridecommandlockouts
\usepackage{amsmath,amssymb,amsfonts}
\usepackage[ruled,vlined]{algorithm2e}
\usepackage{graphicx}
\usepackage{float}
\usepackage{cleveref}
\usepackage[lofdepth,lotdepth]{subfig}
\usepackage[font=footnotesize,labelfont=bf]{caption}
\usepackage{enumitem}
\usepackage{textcomp}
\usepackage{booktabs}
\usepackage{multirow}
\usepackage{xcolor}
\usepackage[square,numbers]{natbib}
\usepackage[export]{adjustbox}
\usepackage{ragged2e}
\usepackage{algorithm2e}
\usepackage{soul}
\usepackage{textcomp}
\usepackage{dblfloatfix}
\usepackage{tikz}

\newcommand*\circled[1]{\tikz[baseline=(char.base)]{
            \node[shape=circle,draw,inner sep=0.8pt, minimum size=2pt] (char) {#1};}}

\def\BibTeX{{\rm B\kern-.05em{\sc i\kern-.025em b}\kern-.08em
    T\kern-.1667em\lower.7ex\hbox{E}\kern-.125emX}}

\usepackage{fancyhdr,lipsum}
\fancypagestyle{firstpage}{
  \fancyhf{}
  \fancyhead[C]{To appear at the IEEE Transactions on Computer-Aided Design of Integrated Circuits and Systems.}
  \fancyfoot[C]{\thepage}
}

\SetAlCapNameFnt{\footnotesize}
\SetAlCapFnt{\footnotesize}

\newcommand{\rpoint}[1]{\circled{{\fontfamily{pcr}\selectfont\footnotesize{#1}}}}

\begin{document}

\title{DESCNet: Developing Efficient Scratchpad Memories for Capsule Network Hardware\\
\vspace*{-5pt}}

\author{
        Alberto Marchisio\textsuperscript{1}, Vojtech Mrazek\textsuperscript{1, 2}, Muhammad Abdullah Hanif\textsuperscript{1}, Muhammad Shafique\textsuperscript{3}\vspace*{-20pt}
\IEEEcompsocitemizethanks{
\IEEEcompsocthanksitem \textsuperscript{1}Technische Universit{\"a}t Wien (TU Wien), Vienna, Austria
\IEEEcompsocthanksitem \textsuperscript{2}Brno University of Technology, Brno, Czech Republic
\IEEEcompsocthanksitem \textsuperscript{3}Division of Engineering, New York University Abu Dhabi, UAE}
\vspace*{-5pt}
}

\maketitle
\thispagestyle{firstpage}
\begin{small}

\begin{abstract}
Deep Neural Networks (DNNs) have been established as the state-of-the-art method for advanced machine learning applications. Recently proposed by the Google Brain's team, the Capsule Networks (CapsNets) have improved the generalization ability, as compared to DNNs, due to their multi-dimensional capsules and preserving the spatial relationship between different objects. However, they pose significantly high computational and memory requirements, making their energy-efficient inference a challenging task. This paper provides, for the first time, an in-depth analysis to highlight the design- and run-time challenges for the (\mbox{on-chip} scratchpad) memories deployed in hardware accelerators executing fast CapsNets inference. To enable an efficient design, we propose an application-specific memory architecture, called DESCNet, which minimizes the \mbox{off-chip} memory accesses, while efficiently feeding the data to the hardware accelerator executing CapsNets inference. We analyze the corresponding \mbox{on-chip} memory requirement, and leverage it to propose a methodology for exploring different scratchpad memory designs and their energy/area trade-offs.
Afterwards, an application-specific \mbox{power-gating} technique for the \mbox{on-chip} scratchpad memory is employed to further reduce its energy consumption, depending upon the mapped dataflow of the CapsNet and the utilization across different operations of its processing. 

We integrated our DESCNet memory design, as well as another state-of-the-art memory design~\cite{Marchisio2019CapsAcc} for comparison studies, with an open-source DNN accelerator executing Google's CapsNet model~\cite{Sabour2017dynamic_routing} for the MNIST dataset. We also enhanced the design to execute the recent deep CapsNet model~\cite{Rajasegaran2019deepcaps} for the CIFAR10 dataset. Note: we use the same benchmarks and test conditions for which these CapsNets have been proposed and evaluated by their respective teams. The complete hardware is synthesized for a 32nm CMOS technology using the ASIC-design flow with Synopsys tools and CACTI-P, and detailed area, performance and power/energy estimation is performed using different configurations. Our results for a selected Pareto-optimal solution demonstrate no performance loss and an energy reduction of 79\% for the complete accelerator, including computational units and memories, when compared to the state-of-the-art design.


\end{abstract}

\vspace*{-8pt}

\end{small}


\section{Introduction}
Deep Neural Networks (DNNs) have shown state-of-the-art accuracy results for various machine learning (ML)-based applications, e.g., image and video processing, automotive, medicine, and finance. 
Recently, Sabour and Hinton et al. from Google Brain~\cite{Sabour2017dynamic_routing} investigated the Capsule Networks (\textit{CapsNets}), an enhanced type of DNNs which has multi-dimensional capsules instead of uni-dimensional neurons (as used in traditional DNNs). 
The ability to encapsulate hierarchical information of different features (position, orientation, scaling) in a single capsule allows to achieve high accuracy in computer vision applications (e.g., MNIST~\cite{LeCun1998MNIST} and Fashion-MNIST~\cite{Xiao2017Fashion-MNIST} datasets, as shown by the Google's team~\cite{Sabour2017dynamic_routing}). 
Recently, Rajasegaran et al.~\cite{Rajasegaran2019deepcaps} proposed a deeper version of the CapsNets (DeepCaps) that performs well on the image classification task, such as the CIFAR10 dataset~\cite{Krizhevsky2009CIFAR}. 
To reduce the energy consumption and latency, many researchers have designed specialized inference accelerators for both DNNs~\cite{Albericio2016Cnvlutin}\cite{Chen2014DaDianNao}\cite{Chen2016Eyeriss}\cite{Han2016EIE}\cite{Google2017TPU}\cite{Lu2017FlexFlow}\cite{Parashar2017SCNN} and CapsNets~\cite{Marchisio2019CapsAcc}, which are typically based on 2D-arrays of multiply-accumulate (MAC) units. 
The work in~\cite{Marchisio2019CapsAcc} mainly focused on the design of a compute-array along with the optimizations for the dynamic routing algorithm. 
\textit{However, the state-of-the-art have not yet investigated the memory architecture design and management for the CapsNet accelerators, which is a crucial component when considering energy reductions of the overall hardware design}. 
Furthermore, memory optimizations for the traditional DNN accelerators do not work efficiently as they do not account for the distinct processing flow and compute patterns of the CapsNets algorithms. 
\textit{This necessitates investigations for specialized memory architectures for the DNN-accelerators executing CapsNets algorithms, while exploiting their unique processing characteristics and memory access patterns to enable high energy efficiency\footnote{Note: assuming a huge \mbox{on-chip} memory is typically not applicable in resource-constrained embedded applications, e.g., deployed in the IoT-edge devices. Therefore, a memory hierarchy system with \mbox{on-chip} and \mbox{off-chip} memories is preferred in this scenario.}, as targeted in this paper. }

\vspace*{1mm}
\textbf{Target Research Problem and Motivational Analysis:} 
Our detailed experimental analysis in Section~\ref{sec:analysis} illustrates that the energy consumption for both the \mbox{on-chip} and \mbox{off-chip} memories contributes to $96\%$ of the total energy consumption of the CapsNet hardware architecture. 
Therefore, it is crucial to 
investigate the energy-efficient design and management of an \mbox{on-chip} memory hierarchy for CapsNets hardware architectures. 
\textit{The key to achieve high energy efficiency is to exploit the application-specific properties of the CapsNets algorithms, that is, the processing behavior of their unique computational blocks, mapped dataflow, and the corresponding memory access patterns.}
Note, the operations and memory access patterns of the CapsNet inference are distinct from those of the traditional DNNs, as we discuss in Section~\ref{sec:analysis}. 
Therefore, the existing memory architectures for the DNN accelerators may not be efficient when executing CapsNets inference. 
Hence, to understand the corresponding design challenges and the optimization potential, 
we perform a detailed analysis of the memory requirements in terms of size, bandwidth and number of accesses for every stage of the CapsNet inference when mapping it to a dedicated CapsNet accelerator (called \textit{CapsAcc})~\cite{Marchisio2019CapsAcc} and to a traditional DNN accelerator like TPU~\cite{Google2017TPU}.

\begin{figure}[h]
    \centering
    \includegraphics[width=\linewidth]{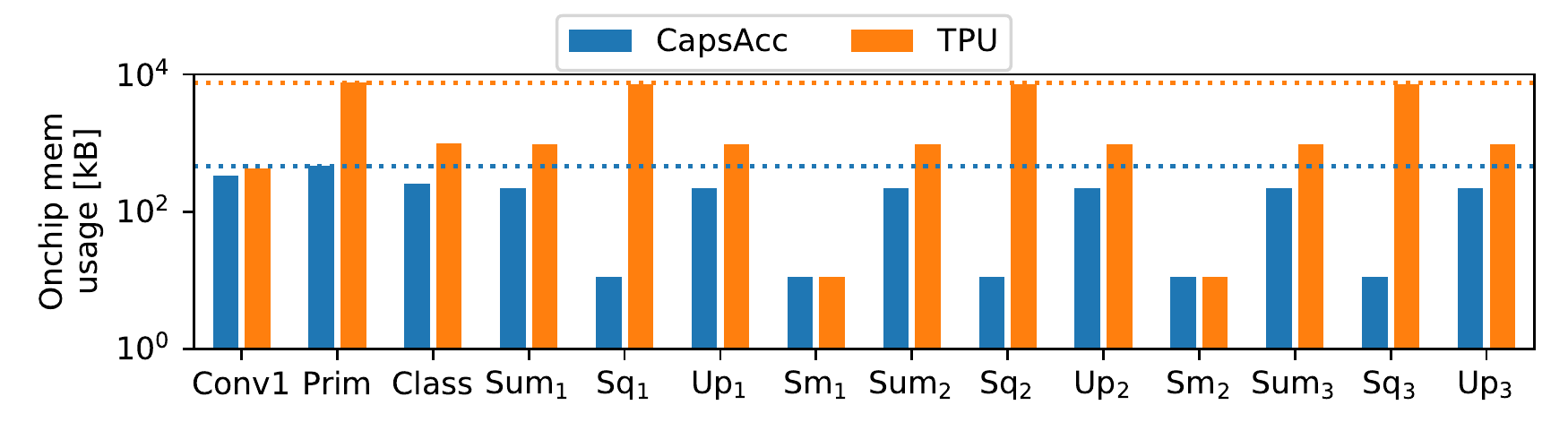}
	\caption{Memory utilization of the Google's CapsNet~\cite{Sabour2017dynamic_routing} inference, mapped on a specialized CapsNet accelerator (\textit{CapsAcc}) \cite{Marchisio2019CapsAcc} and TPU \cite{Google2017TPU}. The bars represent the \mbox{on-chip} memory usage. The dashed lines show the maximum requirement.}
	\label{fig:mem_analysis}
\end{figure}

Fig.~\ref{fig:mem_analysis} shows the \mbox{on-chip} memory utilization in the two accelerator architectures. The dashed lines represent their respective maximum values. 
Note, unique operations (like \textit{ClassCaps}, \textit{Sum}, \textit{Squash}, \textit{Update} and \textit{Softmax}) of the Google's CapsNet~\cite{Sabour2017dynamic_routing} inference mapped onto the \textit{CapsAcc} accelerator of~\cite{Marchisio2019CapsAcc} exhibit different memory utilization profile compared to when mapped to a memory architecture designed for the traditional DNNs like TPU~\cite{Google2017TPU}.
This analysis unleashes the available design and optimization potential for improving the memory energy efficiency when considering a specialized memory architecture for a given CapsNets accelerator.

\textbf{Associated Research Challenges:} 
Traditional memory hierarchies of DNN accelerators \cite{Li2019OnChipMemDNN} are composed of an \mbox{off-chip} DRAM and an \mbox{on-chip} SRAM\footnote{For DNN applications, the \mbox{on-chip} memory is basically a scratchpad memory, and not a traditional cache.}. Though general purpose approaches for memory design exist~\cite{Panda2000OnChipvsOffChip}, achieving a high energy efficiency for CapsNets requires application-specific design and optimizations, as discussed above. 
Such an application-specific design needs to account for the memory access behavior of different processing steps of the CapsNets' algorithms and their respective dataflow on a given CapsNets accelerator, in order to explore the design space of different design parameters of the memory hierarchy (like size, number of banks, partitions, etc.) for multiple levels, which can affect the efficiency of each other. 
For instance, intensive \mbox{on-chip} scratchpad memory (SPM) accesses lead to high energy and performance overhead, requiring a large \mbox{on-chip} memory coupled with CapsNet accelerators supporting efficient data reuse to alleviate the overhead of excessive \mbox{off-chip} memory accesses.
However, a large SPM impacts the chip area and the corresponding leakage power. 
Therefore, besides efficient sizing and partitioning of the SPM, an application-specific \mbox{power-gating}\footnote{Power-gating employs sleep transistors to switch-off the power supply of the unused memory sectors~\cite{Roy2011PowerGating}.} control is also required for further energy reduction under different run-time scenarios of diverse memory usage that are influenced by different processing steps of the executing CapsNet algorithm, while considering the corresponding wakeup overhead\footnote{The \mbox{power-gating} technique comes with the cost of wakeup energy and latency overhead that needs to be considered when taking the \mbox{power-gating} decisions.}.
\textit{Such an application-driven design of memory hierarchy and the power management of the SPM for the CapsNets hardware architectures may bear the potential to provide significant energy savings compared to the traditional memory architectures for DNN accelerators, while keeping a high throughput, as we will demonstrate in this paper.}

\begin{figure}[h]
	\centering
	\includegraphics[width=.8\linewidth]{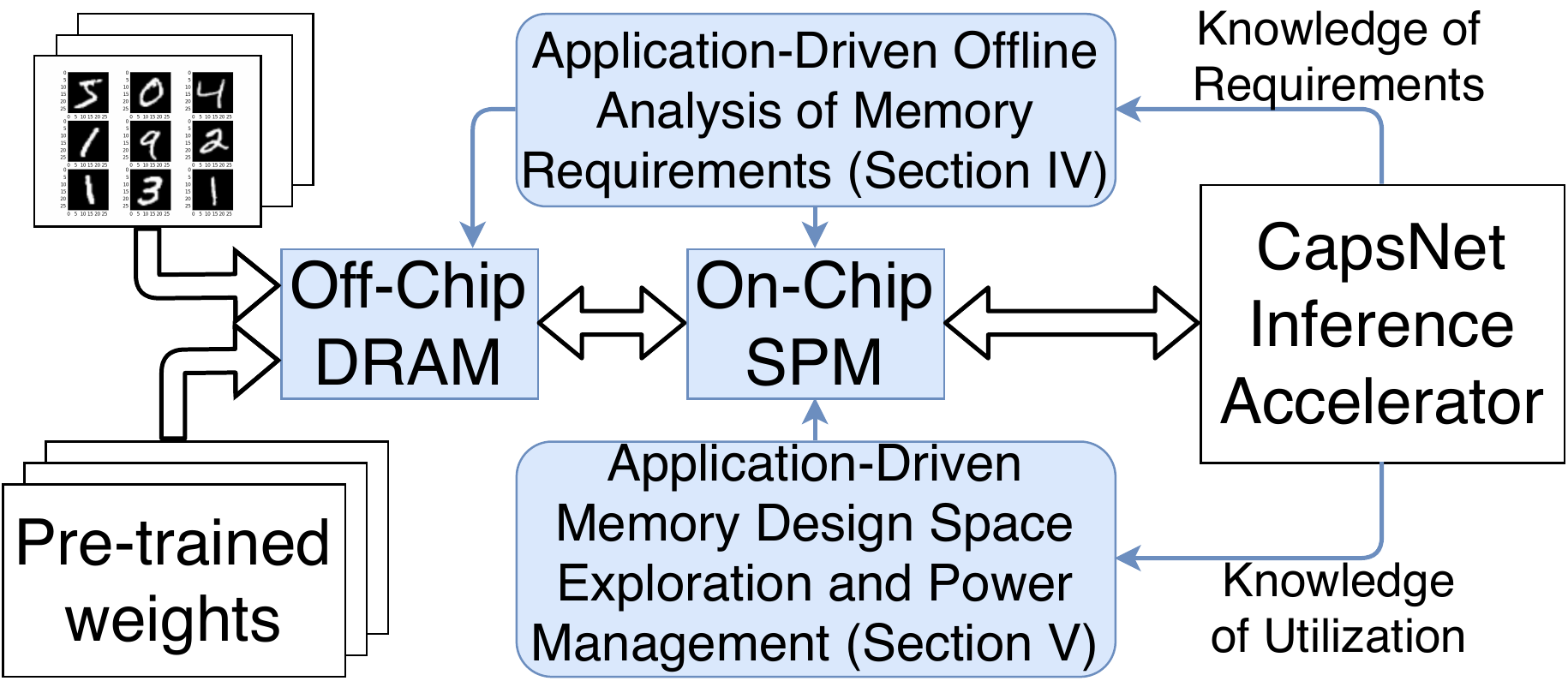}
	\vspace*{2pt}
	\caption{Overview of our \textit{DESCNet} Memory Design Methodology.}
	\label{fig:overview}
\end{figure}

\vspace*{4pt}
\textbf{Our Novel Contributions:} Fig.~\ref{fig:overview} provides an overview of our \textit{DESCNet} memory architecture design methodology, showing the integration of our novel contributions (blue boxes) with a CapsNet accelerator. In a nutshell, we propose:

\begin{enumerate}[leftmargin=*]
    \item \textbf{Memory Analysis of CapsNet Inference (Section~\ref{sec:analysis})} to systematically study the design requirements (size, accesses, etc.), performance and energy consumption, for different operations of the CapsNet~\cite{Sabour2017dynamic_routing} and DeepCaps~\cite{Rajasegaran2019deepcaps} inference.
    
    \item \textbf{\textit{DESCNet}: A Specialized Multi-Banked Scratchpad Memory Architecture for CapsNet Accelerators (Section~\ref{subsec:mem_model})}, which is designed considering the dataflow mapping and the corresponding memory access patterns of different steps of the CapsNet inference. The SPM is partitioned into multiple sectors to support fine-grained sector-level \mbox{power-gating}, thereby providing a higher potential for energy savings under run-time varying memory usage scenarios. Since our SPM supports standard input/output interfaces, it can be coupled with any DNN accelerator that can execute CapsNet inference.
    
    \item \textbf{Design Space Exploration (DSE) (Section~\ref{subsec:DSE_setup})}, which is performed to automatically obtain the Pareto-optimal values of different key parameters of our \textit{DESCNet} memory architecture. It leverages trade-offs between memory, area and energy consumption, while exploiting the distinct processing behavior of different steps of the CapsNet inference.
    
    \item \textbf{Application-Driven Memory Power Management (Section~\ref{subsec:power_management}):} it leverages the processing flow of the CapsNet inference, the architectural parameters of the accelerator, and the interfacing with memory, to devise a sector-level \mbox{power-gating} for reducing the static power.
    
    \item \textbf{Hardware Implementation and Evaluation (Section~\ref{sec:results})} of the complete CapsNet architecture with an integrated \textit{DESCNet} memory in a 32nm CMOS technology using the ASIC-design flow with Synopsys tools and CACTI-P~\cite{Li2011CACTI-P}. We perform area and energy evaluations for 15,233 possible configurations of the \mbox{on-chip} memory architectures for the CapsNet and 215,693 for the DeepCaps, and benchmark them against the state-of-the-art memory design of \cite{Marchisio2019CapsAcc}.
\end{enumerate}

Before proceeding to the technical sections, we present an overview of the CapsNets algorithm and hardware accelerator in \textbf{Section~\ref{sec:background}}, to a level of detail necessary to understand the contributions of this paper. Afterwards, \textbf{Section~\ref{sec:research_questions}} will discuss the required architectural modification towards more practical embedded inference systems, and raise the key research questions to enable efficient memory design and management, that will be addressed by the proposed contributions.

\vspace*{-4pt}
\section{Overview of CapsNets: Algorithm and Hardware Accelerator}
\label{sec:background}

In~\cite{Hinton2011TransformingAutoencoder}, Hinton et al. showed the potential of CapsNets, which exploit the novel concepts of the so-called multi-dimensional \textit{capsules} and the \textit{dynamic routing} algorithm \cite{Sabour2017dynamic_routing} to achieve high accuracy compared to traditional DNNs \cite{Xi2017CapsNet_complex_data}\cite{Mukhometzianov2018CapsNetcomparative}. However, CapsNets are also significantly more complex compared to the traditional DNNs due to their new operations and processing flow, and thereby require new hardware specializations to realize energy-efficient embedded implementations, as showcased in~\cite{Marchisio2019CapsAcc}. More details on the applicability of CapsNets are discussed in Section~\ref{subsec:applicability_caps}.

\vspace*{-8pt}
\subsection{CapsNets vs. DNNs: Differences w.r.t. the Inference}
\label{subsec:capsnet_DNN}

As compared to traditional DNNs, a CapsNet has: 
\begin{itemize}[leftmargin=*]
    \item \textbf{Capsule:} a multi-dimensional neuron, which is able to encapsulate hierarchical information of multiple features (like position, scale, and orientation).
    \item \textbf{Squash Activation Function:} a multi-dimensional non-linear function, which efficiently fits to the prediction vector.
    \item \textbf{Dynamic routing:} an algorithm to learn the connection between two subsequent Capsule layers at run-time during the inference. \textit{It is an iterative algorithm, i.e., it loops over a defined number of routing iterations}.
\end{itemize}

\begin{figure}[h]
	\centering
	\includegraphics[width=.95\linewidth]{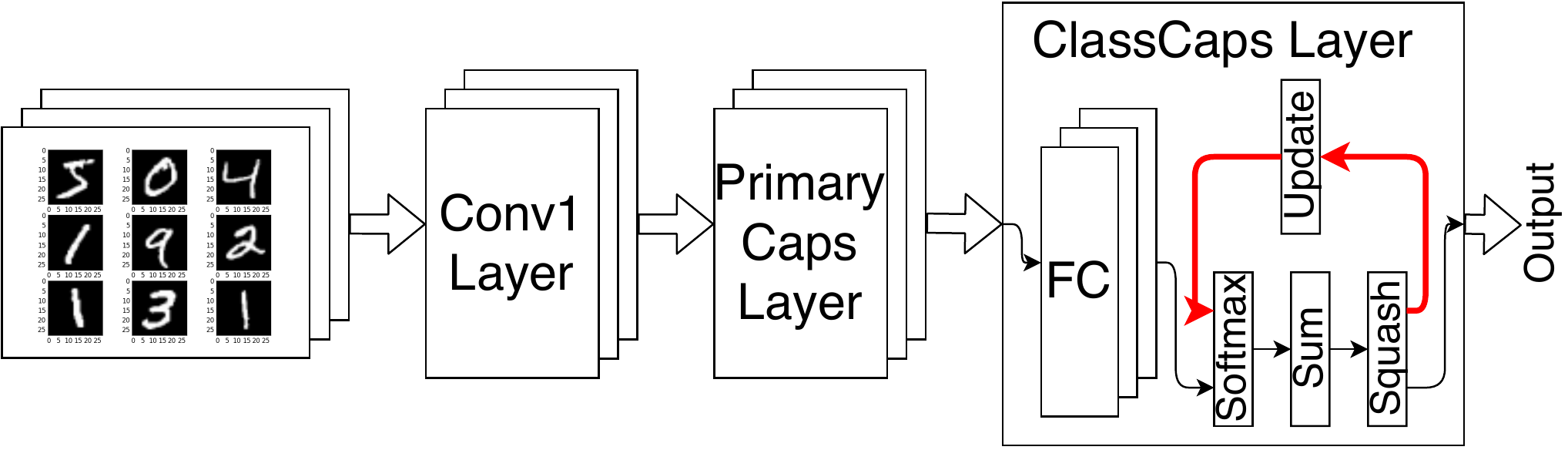}
	\caption{Architecture of the CapsNet for Inference \cite{Sabour2017dynamic_routing}.}
	\label{fig:capsnet_feedback}
\end{figure}

\begin{figure}[h]
	\centering
	\vspace*{-4pt}
	\includegraphics[width=.95\linewidth]{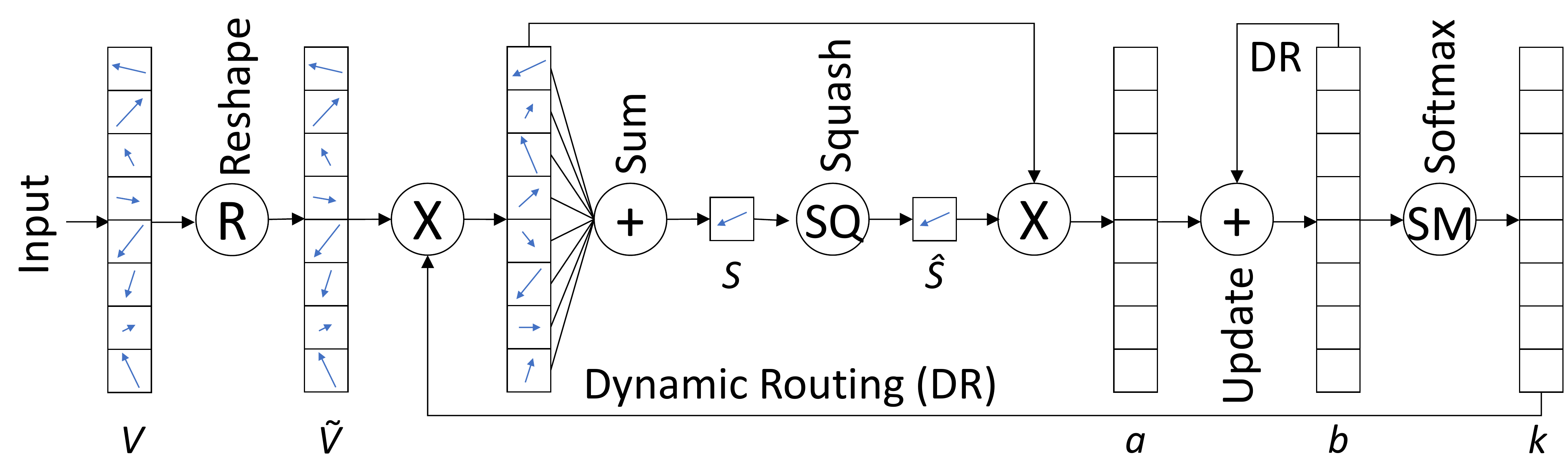}
	\caption{Dynamic routing in CapsNets.}
	\label{fig:routing_by_agreement}
\end{figure}

\textit{The last point is challenging from the hardware design perspective, because, unlike in traditional feed-forward DNN inference, CapsNet inference employs a feedback loop in the \textit{ClassCaps} layer}, as highlighted by the red arrows in Fig.~\ref{fig:capsnet_feedback}. 
Hence, it is more difficult to parallelize and pipeline these computations in the hardware. 
A detailed view of the operations performed in the \textit{ClassCaps} layer, with a focus on the dynamic routing, is depicted in Fig.~\ref{fig:routing_by_agreement}.

\begin{figure}[h]
    \centering
    \includegraphics[width=\linewidth]{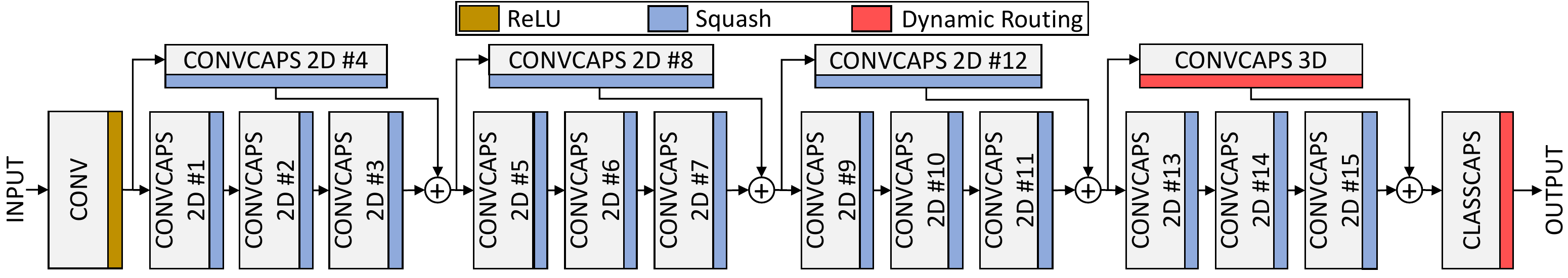}
    \caption{The DeepCaps architecture of~\cite{Rajasegaran2019deepcaps}.}
    \label{fig:deepcaps}
\end{figure}

Recently, an advanced deep CapsNet model, \textit{DeepCaps}~\cite{Rajasegaran2019deepcaps}, has been proposed (see Fig.~\ref{fig:deepcaps}). It introduces skip connections, as well as 2-dimensional (2D) and 3D convolutional layers of capsules (\textit{ConvCaps}). After the first convolutional layer with ReLU activation function, the network features 15 ConvCaps2D layers, with squash activation function. Every three sequential ConvCaps layers have an additional ConvCaps2D layer that operates in parallel. The last parallel ConvCaps layer is 3D and performs the dynamic routing. The output layer of the DeepCaps architecture is a fully-connected capsule layer (\textit{ClassCaps}) with dynamic routing.

\vspace*{-5pt}
\subsection{The CapsNet Accelerator Architecture under Consideration}
\label{subsec:capsnet_accel}

Fig.~\ref{fig:capsacc} shows the architecture of a state-of-the-art hardware accelerator for CapsNet inference called \textit{CapsAcc}~\cite{Marchisio2019CapsAcc}. 
Although our memory designs support standard input/output interfaces such that they can be integrated with any DNN accelerator that can execute CapsNets, we adopt the \textit{CapsAcc} in our evaluation studies because it is the only open-source DNN hardware accelerator executing CapsNets. 
The \textit{CapsAcc} consists of specialized hardware modules like for the Squash function, a 2D MAC-based neural processing (NP) array\footnote{MAC-based NP array architectures have also been proposed for other DNN accelerators like~\cite{Chen2016Eyeriss} and~\cite{Google2017TPU}, but they differ from the \textit{CapsAcc}~\cite{Marchisio2019CapsAcc} in terms of their dataflow, inter-layer transfer of prediction vectors due to the dynamic routing algorithm, and other connecting hardware modules.} 
(16x16 Processing Elements) for efficiently parallelizing the capsule computations, and a specialized dataflow policy for CapsNets. 
The dedicated connections between the accelerator and the memories allow data and weight reuse, properties which are particularly effective for the dynamic routing computation. However, the work of~\cite{Marchisio2019CapsAcc} employs a relatively large \mbox{on-chip} memory\footnote{An implementation with 8MiB \mbox{on-chip} memory is shown with a 16x16 NP array.} to simultaneously store all the memory elements on chip, which can immediately become an impractical solution due to its lack of scalability to large-sized CapsNets and non-applicability in resource-constrained embedded systems. 

\vspace*{-5pt}
\subsection{Applicability of Capsule Networks}
\label{subsec:applicability_caps}

CapsNets represent emerging deep learning algorithms for a wide variety of applications. Initially proposed by the Google Brain's team~\cite{Sabour2017dynamic_routing}\cite{Hinton2011TransformingAutoencoder} for image classification, several architectural variants of Capsule Networks have been proposed~\cite{Rajasegaran2019deepcaps}\cite{Ahmed2019StarCaps}\cite{Hahn2019SelfRoutingCaps}\cite{Tsai2020CapsNetAttentionRouting}\cite{Marchisio2020NASCaps}, as well as CapsNets for other applications, such as action detection~\cite{Duarte2018VideoCapsuleNet}, object segmentation~\cite{LaLonde2018CapsulesSegmentation}, and machine translation~\cite{Wang2019MachineTranslationCapsule}.
Moreover, Saqur et al.~\cite{Saqur2020CapsGAN} proposed to use CapsNets as the discriminator for the Generative Adversarial Networks, while Qin et al.~\cite{Qin2020DetectingAdvCapsule} proposed to employ CapsNets for detecting adversarial images.

\begin{figure}[h]
	\centering
	\includegraphics[width=\linewidth]{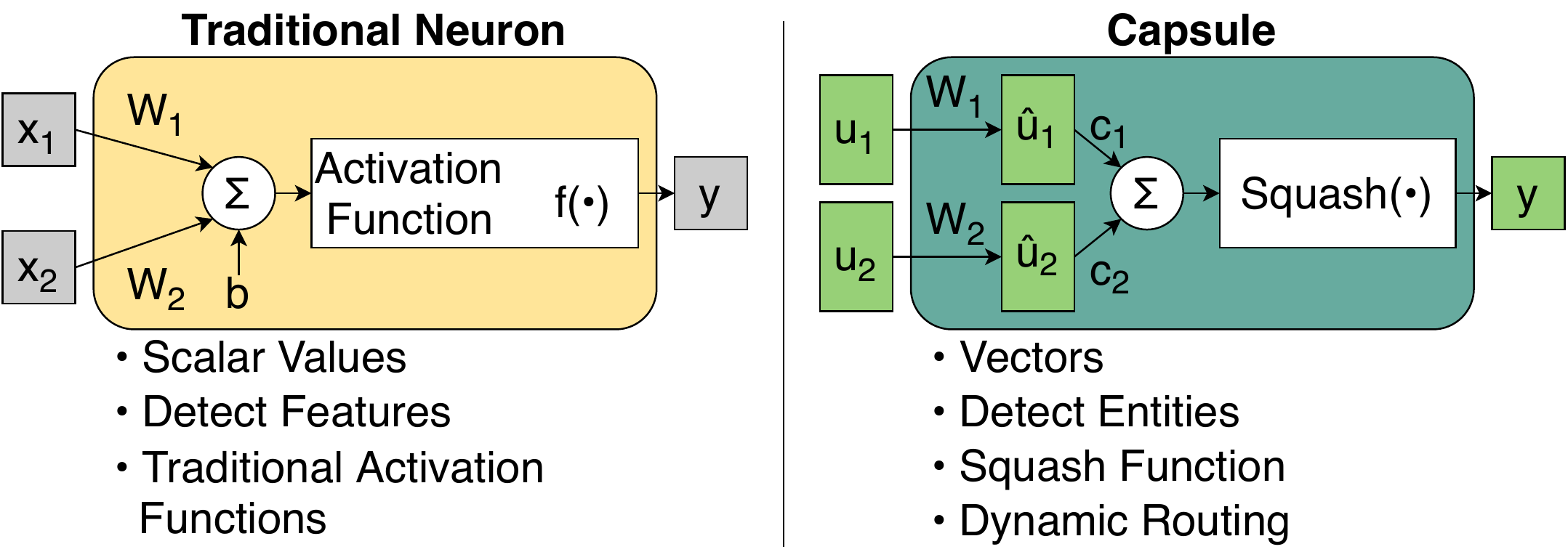}
	\caption{Summary of differences between a traditional neuron and a capsule.}
	\label{fig:cnn_vs_capsnet}
\end{figure}

Overall, the key differences between traditional neurons and capsules are summarized in Fig.~\ref{fig:cnn_vs_capsnet}. While traditional neurons directly multiply and accumulate the scalar input values, the capsules first apply a transformation matrix to the input vectors and then multiply and accumulate the obtained vectors. Moreover, the squash operation replaces traditional activation functions. The dynamic routing algorithm, already shown in Fig.~\ref{fig:routing_by_agreement}, iteratively learns the coupling coefficients, in such a way that the higher level entities can be detected, rather than the simple features detected by traditional neurons.

\begin{figure}[h]
	\centering
	\includegraphics[width=.8\linewidth]{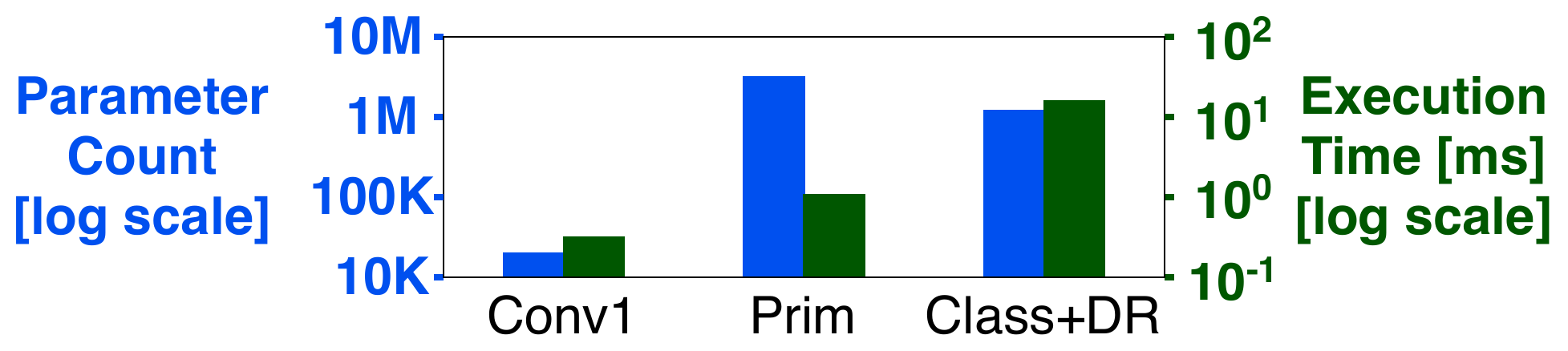}
	\caption{Parameter count vs. execution time of the Google's CapsNet on the Nvidia Ge-Force GTX 1070 GPU.}
	\label{fig:perf_breakdown}
\end{figure}

From the hardware perspective, Marchisio et al.~\cite{Marchisio2019CapsAcc} proposed a CMOS-based hardware architecture for accelerating CapsNets, while Zhang et al.~\cite{Zhang2020PIMCapsNet} designed a Processing-In-Memory architecture for CapsNets. Moreover, the works in~\cite{Marchisio2020ReDCaNe} and~\cite{Marchisio2020QCapsNets} show that CapsNets offer huge potential of energy-efficiency improvements when specialized optimizations such as approximate computing and quantization are applied. However, the key takeaway message from the works in~\cite{Marchisio2019CapsAcc} and~\cite{Zhang2020PIMCapsNet} is that the dynamic routing operations constitute the bottleneck for CapsNet inference, as demonstrated by the analysis of the performance breakdown of the Google's CapsNet executed on the Nvidia Ge-Force GTX 1070 GPU, shown in Fig.~\ref{fig:perf_breakdown}.

\vspace*{-5pt}
\section{Required Architectural Modification and Key Research Questions}
\label{sec:research_questions}

\begin{figure}[h]
\centering
\begin{minipage}[t]{.49\linewidth}
\centering
\subfloat[]{
\includegraphics[width=\linewidth]{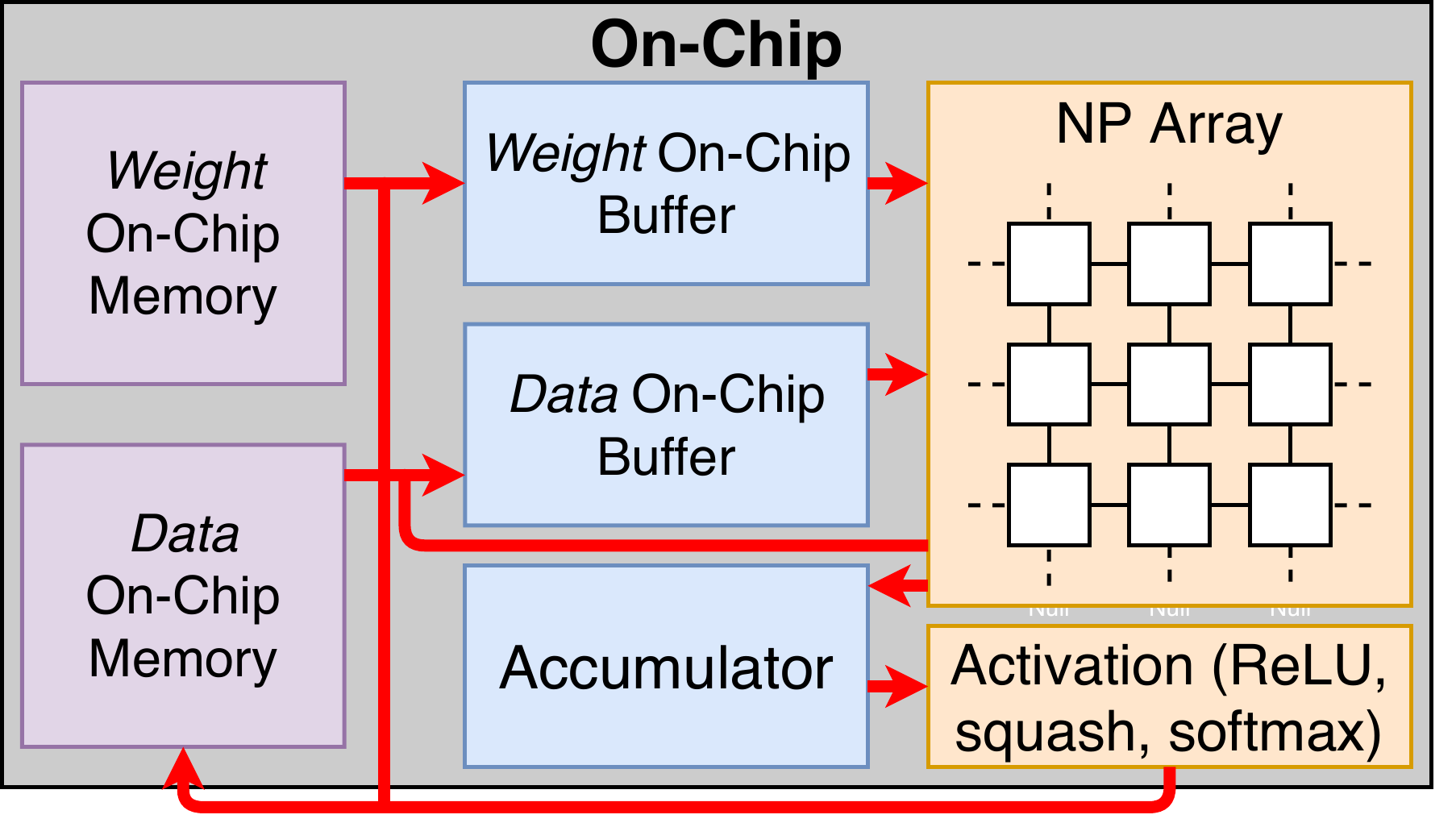}
\label{fig:capsacc}}
\end{minipage}
\hfill
\begin{minipage}[t]{.49\linewidth}
\centering
\subfloat[]{
\includegraphics[width=\linewidth]{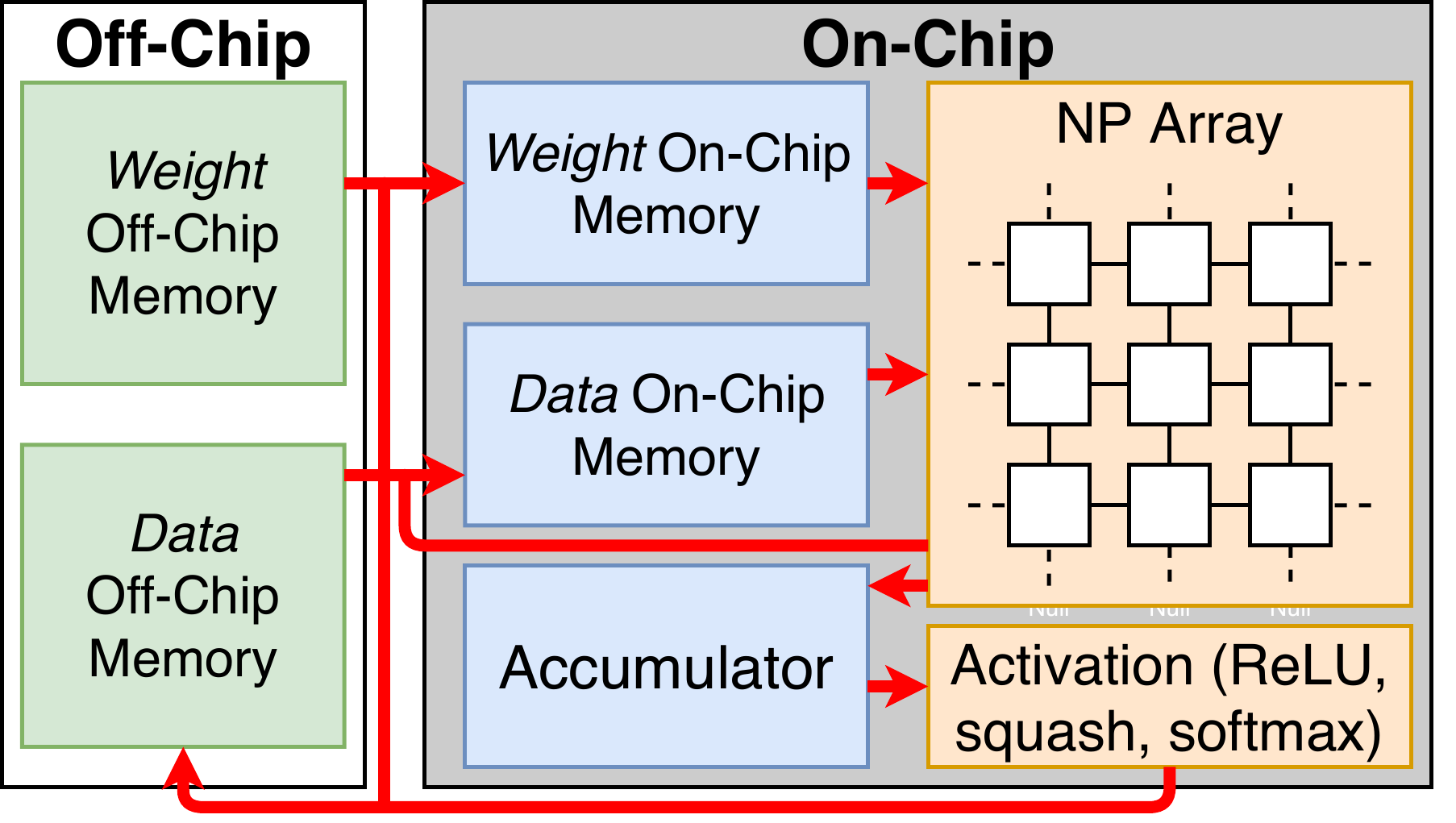}
\label{fig:capsacc_dac}}
\end{minipage}
\caption{Architectural view of the CapsNet inference accelerator. \textbf{(a)} Baseline architecture of~\cite{Marchisio2019CapsAcc}. \textbf{(b)} Modified architecture for this work with \mbox{off-chip} and \mbox{on-chip} memory partitioning, which is more practical for embedded \mbox{implementations.}}
\label{fig:capsacc_architectures}
\end{figure}

To overcome the limitation of the base-architecture of~\cite{Marchisio2019CapsAcc}, and towards real-world embedded implementations, we employ a modified architectural model of CapsNet hardware with a memory hierarchy consisting of an \mbox{on-chip} SPM and an \mbox{off-chip} DRAM, as shown in the blue-colored boxes and green-colored boxes of Fig.~\ref{fig:capsacc_dac}, respectively.

Such a solution can generalize the problem for different applications and more complex CapsNet architectures. 
However, given such a memory hierarchy, the challenge then lies in designing and managing the \mbox{on-chip} memory such that 
(i) the \mbox{off-chip} memory accesses are minimized, 
(ii) the reuse of weights and intermediate data stored in the \mbox{on-chip} memory is exploited at the maximal, and 
(iii) the unnecessary parts of the \mbox{on-chip} memory can be power-gated under scenarios of varying memory accesses without affecting the performance of CapsNets processing. 
These problems have not yet been studied for the CapsNets hardware. \textit{Towards this, we aim at investigating the following key questions when determining the memory sizes and the communication between \mbox{off-chip} and \mbox{on-chip} memories.}
\begin{enumerate}[leftmargin=*]
    \item How to minimize the \mbox{off-chip} memory accesses to reduce the access energy? Every data is read from / written to the \mbox{off-chip} memory only once, while it is used once or multiple times in the \mbox{on-chip} memory.
    \item How to keep the latency and throughput {\it similar/close} to the case of having all the memory \mbox{on-chip}\footnote{This can be guaranteed by prefetching the data to the \mbox{on-chip} memory to mask the \mbox{off-chip} memory latency, assuming that the \mbox{on-chip} memory is large enough to contain all the necessary data.}, i.e., hiding the \mbox{off-chip} latency as much as possible?
    \item How to minimize the \mbox{on-chip} SPM size to reduce the leakage, while it becomes a contradicting design requirement w.r.t. the above-discussed questions 1) and 2)? Hence, what would be the appropriate design trade-offs? 
    \item How can we design efficient \mbox{power-gating} for the \mbox{on-chip} SPM to save the leakage power for the unused sectors?
    \item Can we exploit the unique processing and data reuse behavior of CapsNets to address the above questions and to optimize the corresponding memory access profiles?
\end{enumerate}
Since the above-discussed questions can pose contradicting requirements and constraints, there is a need for an in-depth analysis of the resource requirement and usage patterns of CapsNets processing before taking appropriate design decisions, which we discuss in the following Section~\ref{sec:analysis}.

\vspace*{-5pt}
\section{Resource Analysis of CapsNet Inference}
\label{sec:analysis}

First we investigate the Google's CapsNet architecture of~\cite{Sabour2017dynamic_routing} that performs MNIST~\cite{LeCun1998MNIST} classification, using the architectural organization presented in Fig.~\ref{fig:capsacc_dac}. 
We analyze the performance and the \mbox{on-chip} memory requirements for different operations of the CapsNet inference, showing their \mbox{on-chip} read and write accesses. 
Afterwards, we analyze the DeepCaps~\cite{Rajasegaran2019deepcaps} for the CIFAR10~\cite{Krizhevsky2009CIFAR} classification, showing that, overall, the accumulators have the major contributions in memory usage and accesses. 
Moreover, the energy breakdown analysis highlights the respective contributions of the accelerator and the memories. 
Note that our experiments obtain the same classification accuracies for these CapsNets as they have been proposed by their respective teams, i.e., 99.67\% for the CapsNet on the MNIST dataset and 92.74\% for the DeepCaps on the CIFAR10 dataset.

\vspace*{-8pt}
\subsection{Performance and Memory Analyses for the Google's CapsNet~\cite{Sabour2017dynamic_routing} on the MNIST Dataset}
\label{subsec:perf_mem_breakdown}

\textbf{Memory Usage Analysis:} 
Considering the design options discussed in Section~\ref{subsec:capsnet_accel}, we analyze the \mbox{on-chip} memory requirements for each operation of the CapsNet inference, as shown in Fig.~\ref{fig:mem_analysis}. 
The dashed lines in Fig.~\ref{fig:mem_analysis} represent the maximum values.
The overall size can be determined by the operation that requires the largest amount of memory (i.e., the {\it PrimaryCaps} layer).
For this configuration, the \mbox{on-chip} SPM is composed of the data, weight and accumulator memories. 

\begin{figure}[h]
    \centering
    \vspace*{3pt}
    \includegraphics[width=\linewidth]{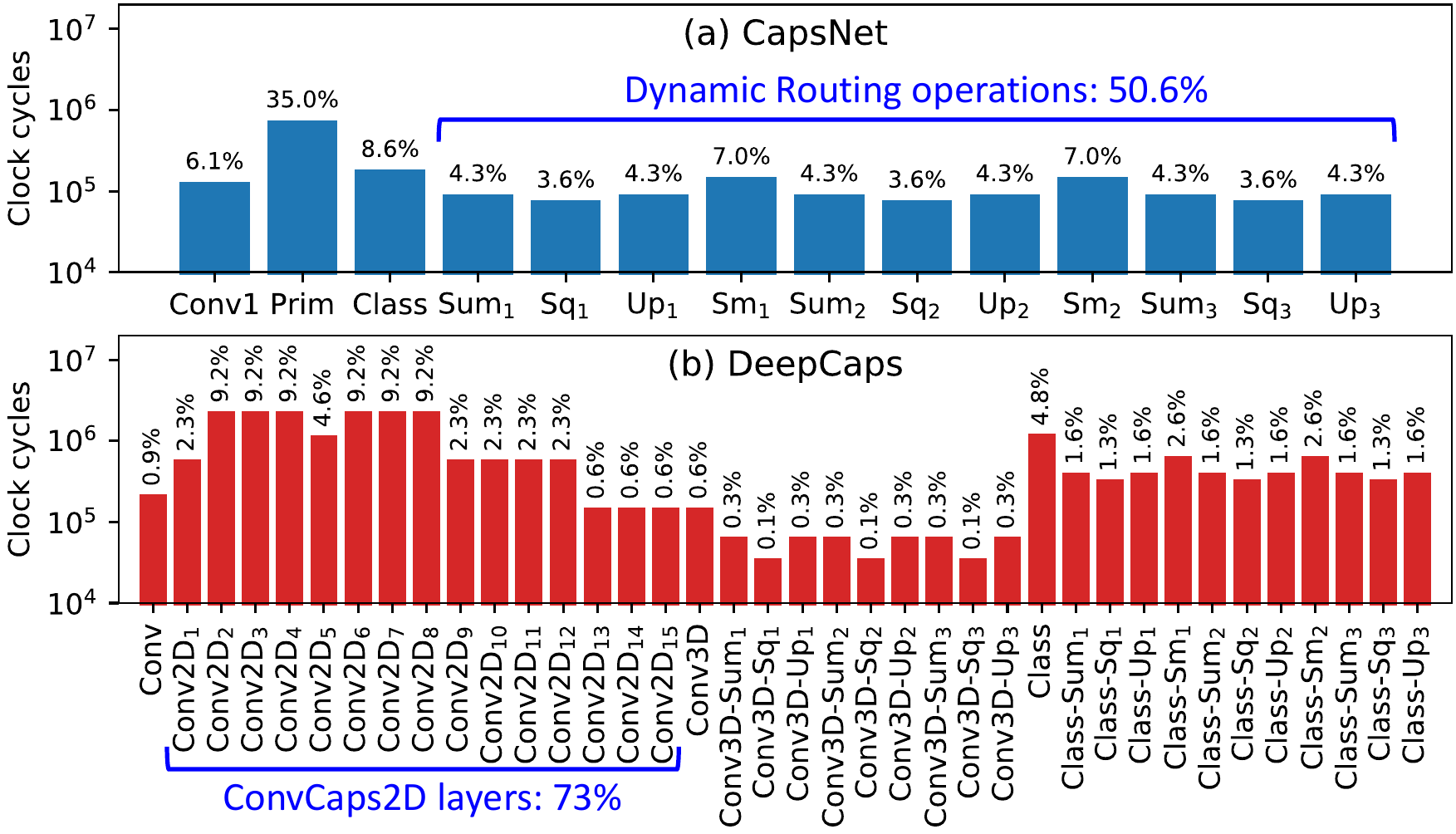}
    \vspace*{-10pt}
    \caption{Clock cycles for different inference operations in (\textbf{a}) CapsNet; (\textbf{b}) DeepCaps.}
    \label{fig:cyc_analysis}
\end{figure}

\vspace*{5pt}
\textbf{Performance Analysis:} 
Fig.~\ref{fig:cyc_analysis}a presents the execution time (i.e., number of clock cycles) of different operations involved in the CapsNet inference. 
Note, the dynamic routing operations contribute for more than half of the execution time of the complete CapsNet inference. 
Overall, the performance is 116 frames-per-second (FPS) for the CapsAcc accelerator, as also reported in the work of~\cite{Marchisio2019CapsAcc}.
If we combine the results of \Cref{fig:mem_analysis,fig:cyc_analysis}a, we notice that, potentially, a significant amount of leakage energy can be saved by the \mbox{power-gating} part of the \mbox{on-chip} memory, when the utilization is below $100\%$. 
We leverage this observation to develop an application-specific power management policy for memories of the CapsNet Accelerators, as discussed in Section~\ref{subsec:power_management}.

\begin{figure}[h]
	\centering
	\includegraphics[width=\linewidth]{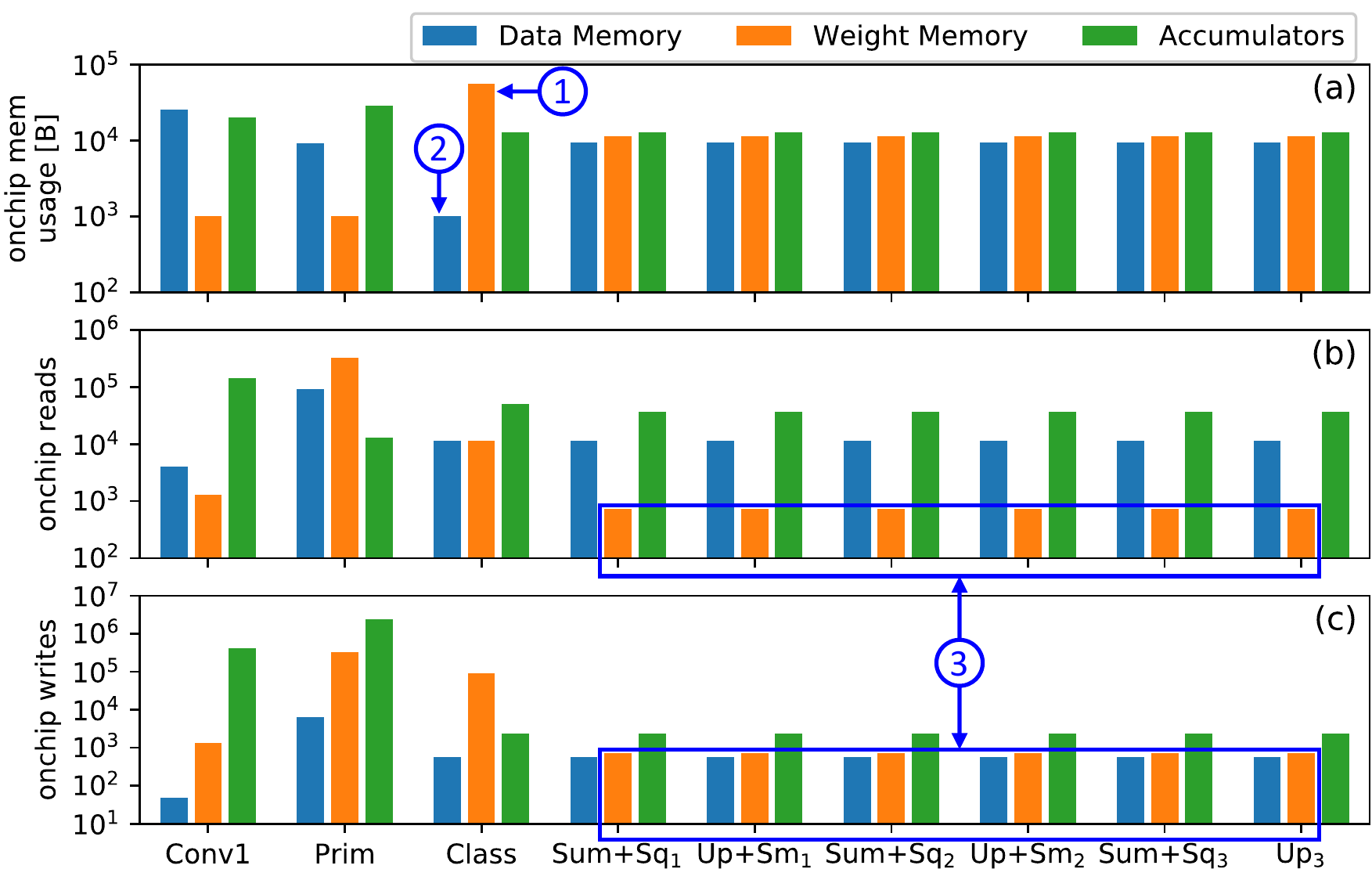}
	\caption{On-chip memory usage, reads and writes of different operations for the CapsNet Inference.}
	\label{fig:rw_caps}
\end{figure}

\vspace*{5pt}
\textbf{Memory Access Analysis:} 
Fig.~\ref{fig:rw_caps}a provides a detailed analysis 
for each memory component (i.e., data memory, weight memory and accumulators), which enables an efficient DSE (Section~\ref{subsec:DSE_setup}) of different architectural parameters of the {\it DESCNet}.
Note, \mbox{handling} different memory components separately may enable efficient power management. 
Figures~\ref{fig:rw_caps}b and~\ref{fig:rw_caps}c show the read and write accesses, respectively, for each operation $i$ of the CapsNet inference, i.e., {\it Convolutional1 (Conv1)}, {\it PrimaryClass (Prim)} and {\it ClassCaps-FullyConnected (Class)}. 
These values are needed to compute the energy consumption of the memories in the subsequent sections. 


For details of the \mbox{off-chip} memory analysis for the Google's CapsNet on the MNIST dataset, we would like to refer to Appendix~\ref{app:offchip}.1.

\vspace*{5pt}
\textbf{From the above analyses, we derive these key observations}: 
\begin{itemize}[leftmargin=*]
    \item For most operations, the accumulator's memory usage is more than the data and weight memories for each operation, because it must store the temporary partial sums of different output feature maps. However, the peak (see pointer \rpoint{1} in Fig.~\ref{fig:rw_caps}) is visible in the weight memory of the \textit{ClassCaps} layer, which is fully-connected.
    \item Data and weight memory requirements vary significantly across different operations.
    \item In the first two layers, the weight memory usage is quite low as compared to the other stages, because the architecture can efficiently employ weight reuse for the convolutions.
    \item In the {\it ClassCaps} layer the data memory usage is low (see pointer \rpoint{2} in Fig.~\ref{fig:rw_caps}), because the corresponding data reuse is efficient.
    \item Weight reuse is relatively more efficient in the last six operations (dynamic routing), as compared to the first three (see pointer \rpoint{3} in Fig.~\ref{fig:rw_caps}).
    \item During the dynamic routing, the \mbox{off-chip} memory is not accessed, except for read accesses in the first operation and write accesses in the last one (see pointer \rpoint{4} in Fig.~\ref{fig:offc_caps}), due to the efficient data and weight reuse in these operations.
\end{itemize}

\vspace*{-8pt}
\subsection{Performance and Memory Analyses for the DeepCaps~\cite{Rajasegaran2019deepcaps} on the CIFAR10 Dataset}
\label{subsec:mem_usage_deepcaps}

Similar analyses have also been carried on for a deeper and more complex capsule network such as the DeepCaps~\cite{Rajasegaran2019deepcaps}. 
The performance in terms of clock cycles is shown in Fig.~\ref{fig:cyc_analysis}b, and overall it is 9.7 FPS. 
Compared to the CapsNet, the DeepCaps shows a more distributed partition. 
Overall, we can notice that the most time-consuming operations are in the \textit{ConvCaps2D} layers, which contribute for 73\% of the execution time of the complete DeepCaps inference.

\begin{figure}[h]
	\centering
	\includegraphics[width=\linewidth]{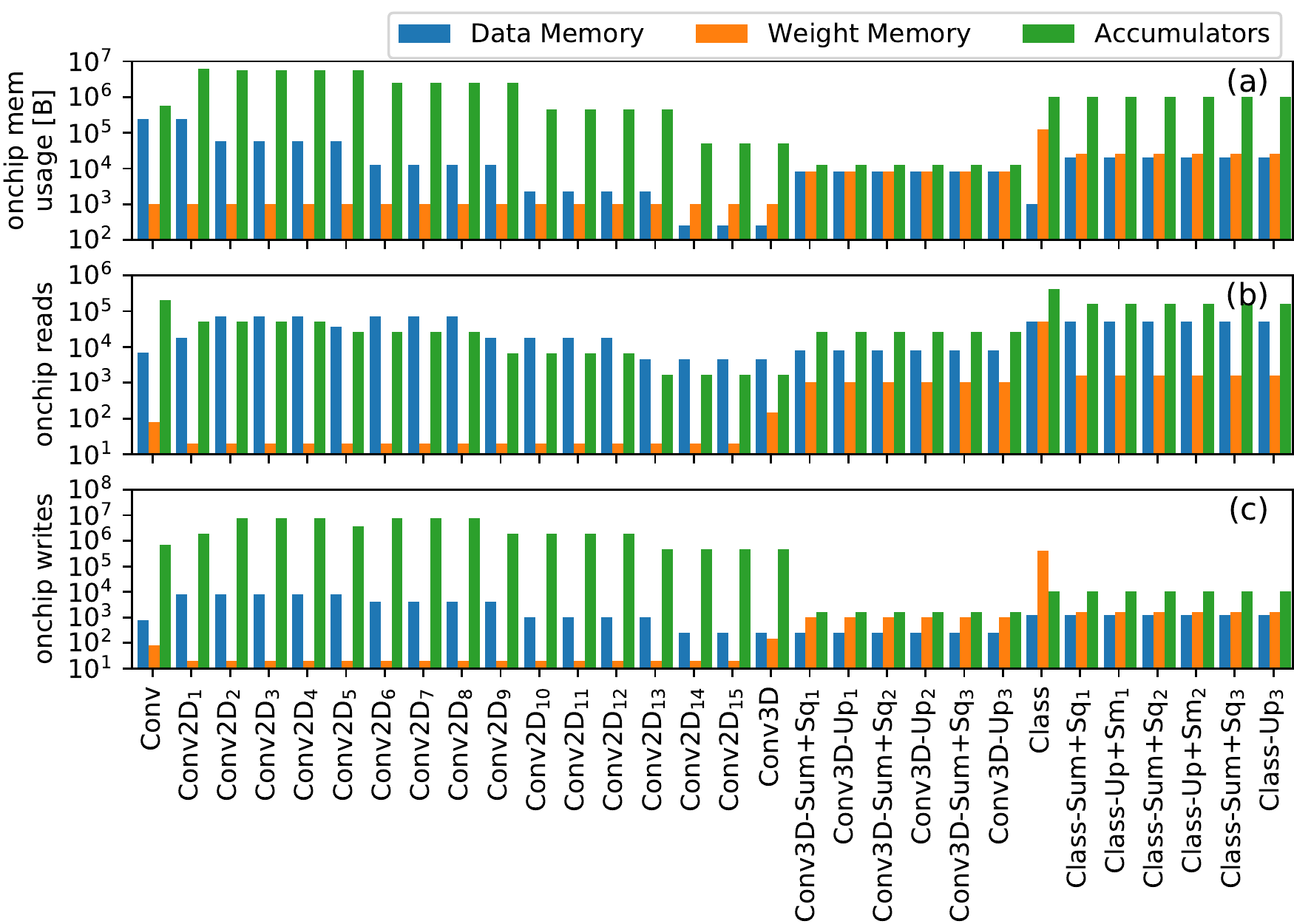}
	\caption{On-chip memory usage, reads and writes of different operations for the DeepCaps inference.}
	\label{fig:rw_deepcaps}
\end{figure}

The \mbox{on-chip} memory usage, reads and writes are shown in Fig.~\ref{fig:rw_deepcaps}. 
Similar to the case of the CapsNet, the accumulator's usage is higher than the data and weight memories. 
Moreover, the usage and accesses for the weight memory are low in the convolutional layers, but higher for the dynamic routing operations.

For the \mbox{off-chip} accesses for the DeepCaps on the CIFAR10 dataset, we refer to Appendix~\ref{app:offchip}.2.

\vspace*{-8pt}
\subsection{Energy Breakdown Analysis}
\label{subsec:energy_breakdown}

To compute the energy consumption of the complete architecture, we develop the following two different versions.

\begin{enumerate}[label=(\alph*), leftmargin=5.5mm]
    \item Fig.~\ref{fig:capsacc}  \cite{Marchisio2019CapsAcc}: 
    an accelerator (composed of NP array, activation unit and control unit), \mbox{on-chip} {SPM} buffers (data
    , weight 
    and accumulator's memory), and an \mbox{on-chip} SPM (for data and weights). 
    The total \mbox{on-chip} memory is of $8MiB$. 
    
    \item Fig.~\ref{fig:capsacc_dac}: an accelerator with the same composition as above, but different architectures of \mbox{on-chip} and \mbox{off-chip} memories. The sizes are derived from the analyses of Section~\ref{subsec:perf_mem_breakdown}.
\end{enumerate}

\begin{figure}[h]
\centering
\vspace*{0pt}
\begin{minipage}[t]{.46\linewidth}
\vspace*{0mm}
\subfloat[]{
\includegraphics[width=\linewidth]{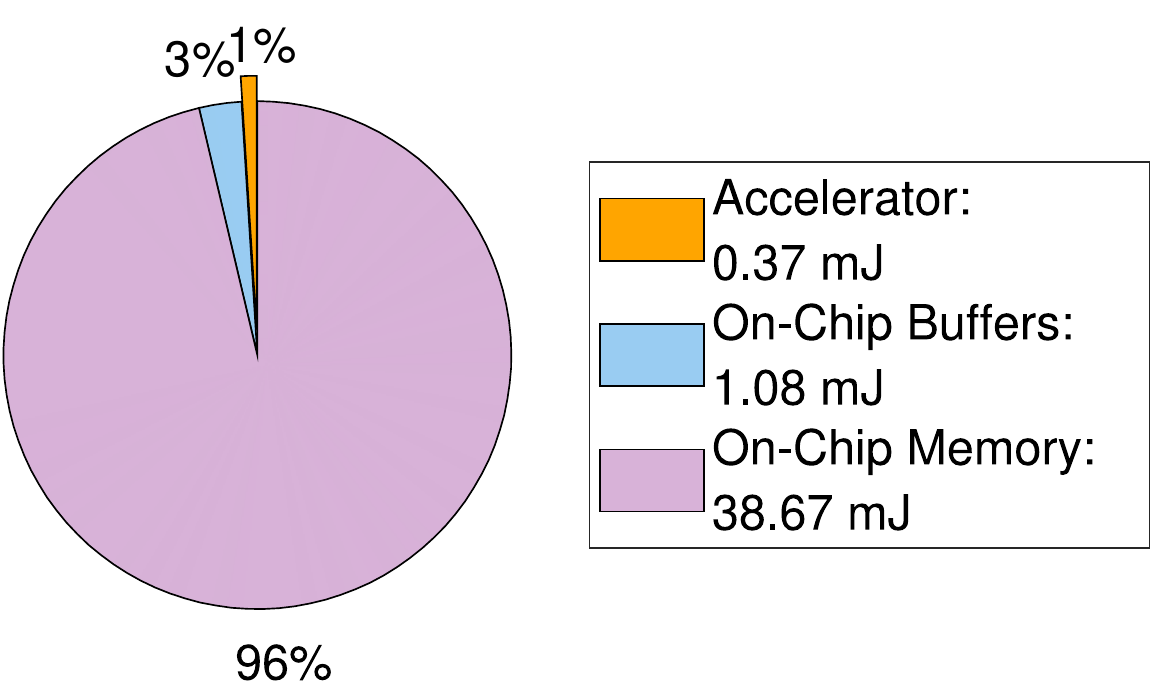}
\label{fig:energy_breakdown_sram}}
\end{minipage}
\hfill
\begin{minipage}[t]{.52\linewidth}
\vspace*{2pt}
\subfloat[]{
\includegraphics[width=\linewidth]{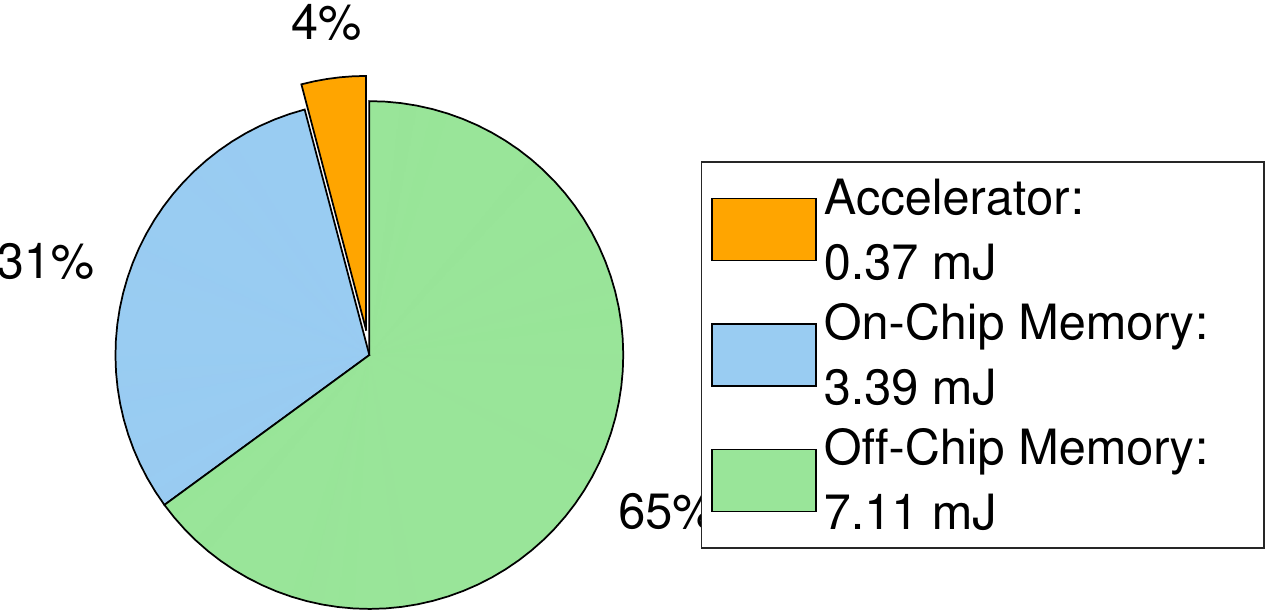}
\vspace{10mm}
\label{fig:energy_breakdown_dram}}
\end{minipage}
\caption{Energy breakdown of different components of the CapsNet Inference Architecture: considering \textbf{(a)} all \mbox{on-chip}, as employed in~\cite{Marchisio2019CapsAcc} and \textbf{(b)} a memory hierarchy composed of \mbox{on-chip} and \mbox{off-chip} memories as shown in Fig.~\ref{fig:capsacc_dac}.}
\label{fig:energy_breakdown}
\end{figure}

The energy breakdown is shown in Fig.~\ref{fig:energy_breakdown}. 
The results are obtained by synthesizing the DNN accelerator of~\cite{Marchisio2019CapsAcc}, executing the Google's CapsNet~\cite{Sabour2017dynamic_routing} for the MNIST dataset, in a 32nm CMOS technology, while the \mbox{on-chip} and \mbox{off-chip} memory values are obtained using the CACTI-P tool \cite{Li2011CACTI-P} with the compatible technology parameters, as it is well-adopted by the memory community.

Our analysis shows that by designing a different memory hierarchy we can already save $73\%$ of the total energy, as compared to the state-of-the-art architecture in~\cite{Marchisio2019CapsAcc}. This can be attributed to the significantly reduced leakage energy due to the lower \mbox{on-chip} memory size. 
Moreover, the \mbox{on-chip} memory consumes \textit{$31\%$ of the total energy}, which corresponds to \textit{the $90\%$ of the \mbox{on-chip} energy} (i..e., accelerator and SPM). 
Hence, an application-driven memory power management (see Section~\ref{subsec:power_management}) can have a significant impact on the overall \mbox{on-chip} energy savings. 

Note: the DeepCaps' execution for the CIFAR10 dataset \mbox{cannot} be supported on the original baseline CapsAcc~\cite{Marchisio2019CapsAcc} due to its \mbox{memory} requirements (i.e. DeepCaps does not fit in the $8MiB$ memory of~\cite{Marchisio2019CapsAcc}). Hence, a comparison similar to the results of Fig.~\ref{fig:energy_breakdown} for executing the DeepCaps is not feasible. However, as we will demonstrate later in the paper, our proposed DESCNet \mbox{memory} \mbox{architecture} enables the deployment of DeepCaps with low \mbox{hardware} resources.

\vspace*{-5pt}
\subsection{Summary of the Key Observations from our Analyses}
\label{subsec:key_observations}

Summarizing our analyses of~\Cref{subsec:perf_mem_breakdown,subsec:energy_breakdown}, the following key observations can be leveraged to design an efficient memory sub-system for the CapsNet hardware.

\begin{itemize}[leftmargin=*]
\item Most of the energy is consumed by the (\mbox{on-chip} and \mbox{off-chip}) memory, as compared to the computational array.
\item An application-driven memory hierarchy, composed of an \mbox{on-chip} SPM and an \mbox{off-chip} DRAM, can save up to $73\%$ of the energy on the Google's CapsNet model, without compromising the throughput\footnote{The same throughput is guaranteed by prefetching the data for the next operation, in an interleaved fashion with the processing of the current operation. Therefore, the \mbox{off-chip} memory latency can be hidden.}, as compared to having a fully \mbox{on-chip} memory organization.
\item The utilization of the \mbox{on-chip} memory is variable, depending upon the operation of the CapsNet inference. Thus, applying \mbox{power-gating} to the non-utilized sectors can further reduce the energy consumption.
\item Partitioning the SPM into separate components (for data, weight and accumulator) can be beneficial for storing values and efficiently feeding them to the accelerator.
\end{itemize}

\vspace*{-5pt}
\section{DESCNet: Scratchpad Memory Design}
\label{sec:design}

\subsection{DESCNet Memory Architecture}
\label{subsec:mem_model}

\begin{figure}[h]
	\centering
	\includegraphics[width=.95\linewidth]{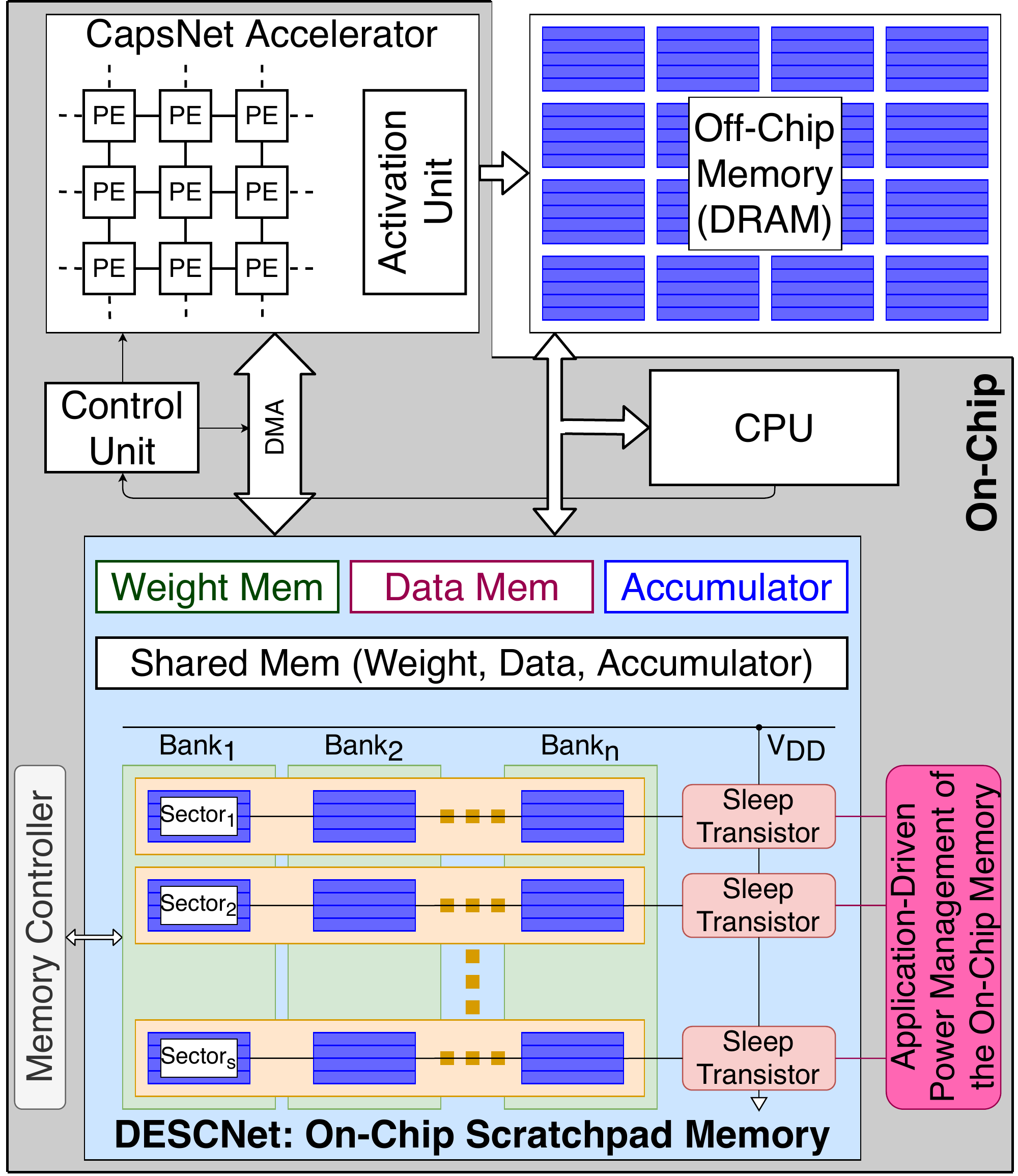}
	\caption{Architectural view of the complete CapsNet architecture, with a focus on our \textit{DESCNet} SPM.}
	\label{fig:memory_model}
	\vspace*{0mm}
\end{figure}

The architecture of our {\it DESCNet} is depicted in Fig.~\ref{fig:memory_model}. It is 
connected to the CapsNet accelerator and to the \mbox{off-chip} memory through dedicated bus lines. The SPM is partitioned into $B$ banks, where each bank consists of $SC$ number of equally-sized sectors. 
All the sectors with the same index across different banks are connected through a \mbox{power-gating} circuitry (implemented with sleep transistors\footnote{i.e., $SC$ sectors, one for each bank, have the same sleep signal. In this work, we only consider $ON$ and $OFF$ modes as they came out to be beneficial in our designs, and we did not need state-retentive modes.}) to support an efficient sector-level power-management control, at the cost of some area overhead. 
Our application-driven memory power management unit determines the appropriate control signals (i.e., \textit{ON} $\leftrightarrow$ \textit{OFF}\footnote{Once the execution of the current operation is completed, the values in the SPM are no more utilized in the next operations. Hence, in our designs, we consider a deep-sleep $OFF$ state, which is non-retentive.}) for the sleep transistors. 
The transitions between sleep modes come at the cost of a certain wakeup energy and latency overhead, that needs to be amortized by the leakage energy savings which depends upon the sleep duration and the number of sectors in the sleep-mode. 
Note, our memory model can be generalized for different memory organizations supporting different sizes and level of parallelism, including multi-port memories.
\textit{Towards this, we study the following three different design options.}

\begin{figure*}[h]
\centering
\vspace*{0mm}
\begin{minipage}[t]{.30\linewidth}
\vspace*{9mm}
\subfloat[]{\adjustbox{valign=T}{
\includegraphics[width=.95\linewidth]{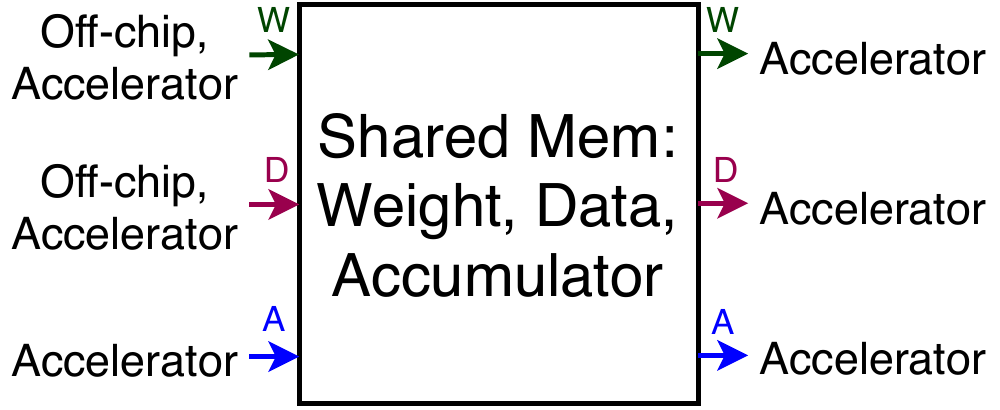}
\label{fig:multiport_mem}}}
\end{minipage}
\hfill
\begin{minipage}[t]{.28\linewidth}
\vspace*{10mm}
\subfloat[]{\adjustbox{valign=T}{
\includegraphics[width=.95\linewidth]{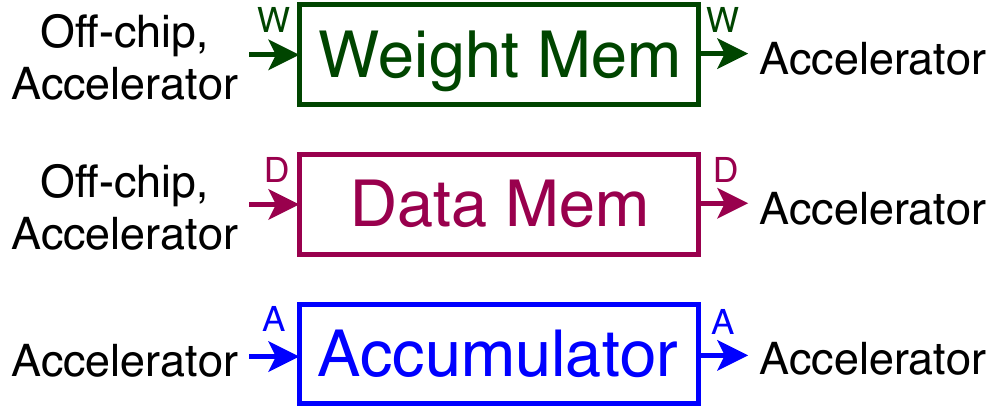}
\label{fig:separate_element}}}
\end{minipage}
\hfill
\begin{minipage}[t]{.38\linewidth}
\vspace*{0mm}
\subfloat[]{
\includegraphics[width=\linewidth]{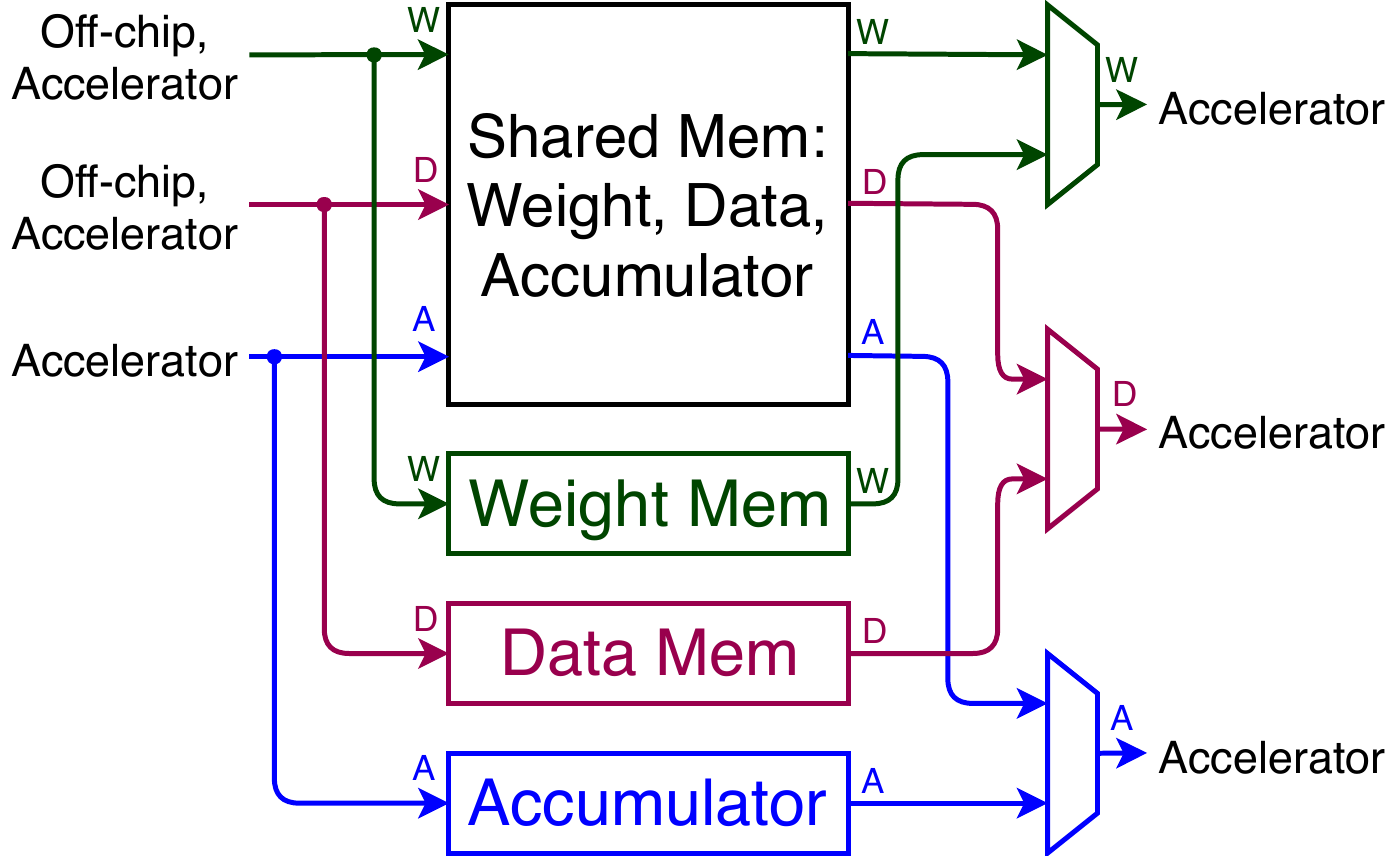}
\label{fig:hybrid_mem}}
\end{minipage}
\caption{Different Architectural Design Options of the On-Chip SPM of the CapsNet Accelerator that are evaluated in our Application-Driven Memory DSE: \textbf{(a)} Shared Multi-Port Memory. \textbf{(b)} Separated Memory. \textbf{(c)} Hybrid Memory (Shared and Separated).}
\label{fig:memory_architectures}
\vspace*{0mm}
\end{figure*}

\vspace*{5pt}
\begin{enumerate}[label=(\alph*),leftmargin=5.5mm]
    \item \textbf{Fig.~\ref{fig:multiport_mem} - Shared Multi-Port Memory (SMP):} a shared \mbox{on-chip} memory 
    with 3 ports for accessing the weights, input data and accumulator's storage in parallel.
    \item \textbf{Fig.~\ref{fig:separate_element} - Separated Memory (SEP):} weights, input data, and the accumulator's partial sums are stored in separate \mbox{on-chip} memories.
    \item \textbf{Fig.~\ref{fig:hybrid_mem} - Hybrid Memory (HY):} a combination of the above two design options, i.e., an \textit{SMP} coupled with a \textit{SEP} memory.
\end{enumerate}


\begin{figure}[h]
	\centering
	\includegraphics[width=.75\linewidth]{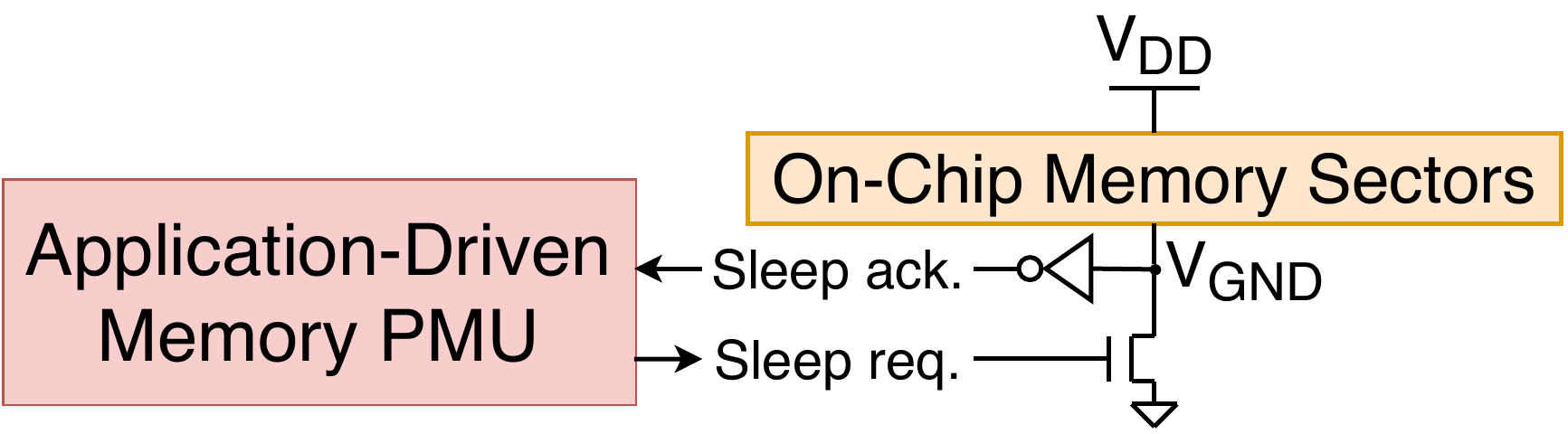}
	\caption{Circuit-level schematic of the \mbox{power-gating} circuit, using a footer sleep transistor connected to the PMU.}
	\label{fig:sleep_circuit}
\end{figure}

\vspace*{-8pt}
\subsection{Application-Driven Memory Power Management}
\label{subsec:power_management}

\begin{figure}[h]
	\centering
	\vspace*{3pt}
	\includegraphics[width=\linewidth]{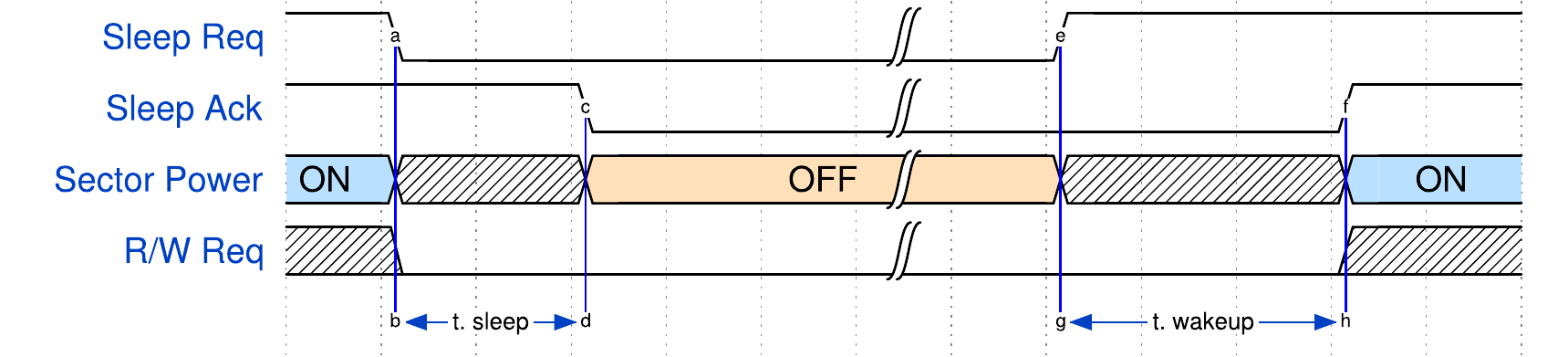}
	\caption{Timing diagram of a complete sleep cycle of a sector.}
	\label{fig:sleep_cycle}
\end{figure}

Our application-driven memory PMU 
determines the sleep signals 
according to the utilization profile of the memory, as observed in Figures~\ref{fig:rw_caps}a and~\ref{fig:rw_deepcaps}a. 
A simple schematic showing how a sleep transistor is connected to its memory sectors is depicted in Fig.~\ref{fig:sleep_circuit}. 
The sleep request is followed by the acknowledge signal, forming a 2-way handshake protocol. 
The timing diagram of a complete sleep cycle (\textit{ON} $\rightarrow$ \textit{OFF} $\rightarrow$ \textit{ON}) is shown in Fig.~\ref{fig:sleep_cycle}. 
When exploiting the application-specific knowledge, it is known from the analysis presented in Section~\ref{sec:analysis} which sectors need to be activated during the execution of different operations. 
Hence, \textit{the wakeup latency overhead is transparently masked, i.e., the required sectors are pre-activated in advance, in such a way that they are active when needed. }


\vspace*{-8pt}
\subsection{Application-Driven Design Space Exploration (DSE) of the DESCNet Memory Designs}
\label{subsec:DSE_setup}

Considering the above memory models, we now determine their organization, sizes, the number of banks ($B$), and the number of sectors-per-bank ($SC$) systematically through a DSE methodology.  We explore different configurations of the memory architecture and evaluate their area and energy consumption. Different levels of abstraction of application-driven knowledge (i.e., architecture and utilization profiles specific to CapsNets) are employed. In the subsequent equations and algorithms, we adopt the following notations:

\begin{itemize}[leftmargin=*]
    \item $i$: index of the operations of the CapsNet inference.
    \item $D_i,\ W_i,\ A_i$: operation-wise memory usage of data memory, weight memory and accumulators, according to the analyses shown in Figures~\ref{fig:rw_caps}a and~\ref{fig:rw_deepcaps}a.
    \item $SZ_{\{S,D,W,A\}},\ SC_{\{S,D,W,A\}},\ B_{\{S,D,W,A\}}$: size, number of sectors, number of banks of \{shared memory, data memory, weight memory, accumulators\}, respectively.
    \item $\sigma(s)$: pool of available numbers of memory sectors for \mbox{power-gating}, given the memory size $s$, which are all the power of two values in the range $\left[ 2, \frac{s}{128} \right]$\footnote{This value is due to a limitation of the CACTI-P~\cite{Li2011CACTI-P} tool, which sets the limit for the ratio between memory size and sector size to be at least 128.}.
\end{itemize}

For all the memory designs, without loss of generality, the number of banks is chosen to be $B_D = B_W = B_A = B_S = 16$, as it corresponds to the number of rows and columns of the NP array of the \textit{CapsAcc} architecture. To facilitate efficient data feeding to the accelerator, this parameter is not changed in our DSE. 

For the \textit{SMP} design, the size of the shared memory is derived from the operation-wise maximum memory usage scenario\footnote{Since only a finite values of memory sizes are acceptable, when computing the $\max_i$ function, the memory size becomes the lowest acceptable size that is greater than or equal to the operation-wise maximum, and vice-versa for the $\min_i$.}, as shown in Equation~\ref{eq:size_SMP}.

\vspace*{-2pt}
\begin{equation}
\vspace*{0pt}
    \verb|SMP:| SZ_S = \max_i (D_i + W_i + A_i)
    \label{eq:size_SMP}
\end{equation}

For the \textit{SEP} design, the sizes of the data memory, weight memory and accumulator are set as in Equation~\ref{eq:size_SEP}, based on the operation-wise maximum memory usage of the separated components.

\vspace*{-2pt}
\begin{equation}
\vspace*{0pt}
    \verb|SEP:| 
    \begin{cases}
    SZ_D = \max_i (D_i)\\
    SZ_W = \max_i (W_i)\\
    SZ_A = \max_i (A_i)
    \end{cases}
    \label{eq:size_SEP}
\end{equation}

For the \textit{HY} design, sizing the memories becomes a more complicated challenge. 
Within the range allowed by the Algorithm~\ref{alg:hybrid_size}, the memory sizes considered in our design space have power-of-two values, with the addition of four randomly selected memory sizes (that are 25 kiB, 108 kiB, 450 kiB, and 460 kiB), to have more fine-grained results in the low-sized range. 
For every possible sizes of data memory, weight memory and accumulators, the size of the shared memory is computed as the operation-wise worst-case that still guarantees the minimum memory usage required by each operation.

\begin{algorithm}[h]
\vspace*{4pt}
\begin{small}
\SetAlgoLined
\SetKwInOut{Input}{Inputs}
\SetKwInOut{Output}{Outputs}
\Input{Operation-wise memory usage $D_i,\ W_i,\ A_i$. 
    }
\Output{Hybrid memory sizes $SZ_S,\ SZ_D,\ SZ_W,\ SZ_A$.}
$ret \leftarrow \{\}$\;
\For{$sz_d\gets \min_i D_i$ \KwTo $\max_i D_i$}{
\For{$sz_w\gets \min_i W_i$ \KwTo $\max_i W_i$}{
\For{$sz_a\gets \min_i A_i$ \KwTo $\max_i A_i$}{
$sz_s \gets \max_i (\max(0, D_i - sz_d) +$
$\max(0, W_i - sz_w) + \max(0, A_i - sz_a))$\;
$ret \leftarrow ret \cup \{(sz_s, sz_d, sz_w, sz_a)\}$\;
}
}
}
\Return $ret$\;
\end{small}
\caption{Exploration of hybrid memory sizes.}
\label{alg:hybrid_size}
\end{algorithm}

After finding the appropriate memory sizes, the \mbox{power-gating} technique can be applied. 
It directly affects the number of sectors in the memory designs, since a sleep transistor is connected to each sector to switch $ON$ or $OFF$ the whole sector. Hence, for the memory designs where the \mbox{power-gating} is not supported, the number of sectors is 1. 
When the \mbox{power-gating} is supported, the choice of number of sectors directly influences the trade-off between the reduction in the static power and the overhead of the \mbox{power-gating} circuitry overhead. 
Towards this, Algorithm~\ref{alg:sector_number} describes all the combinations of valid number of sectors allowed by the function $\sigma(s)$ that are explored. 

\begin{algorithm}[h]
\vspace*{4pt}
\begin{small}
\SetAlgoLined
\SetKwInOut{Input}{Inputs}
\SetKwInOut{Output}{Outputs}
\Input{Memory sizes $sz_s,\ sz_d,\ sz_w,\ sz_a$}
\Output{Number of sectors $SC_s,\ SC_d,\ SC_W,\ SC_A$}
$ret \leftarrow \{\}$\;
\For{$sc_s\in \sigma(sz_s)$}{
\For{$sc_d\in \sigma(sz_d)$}{
\For{$sc_w\in \sigma(sz_w)$}{
\For{$sc_a\in \sigma(sz_a)$}{
$ret \leftarrow ret \cup \{(sc_s, sc_d, sc_w, sc_a)\}$;
}
}
}
}
\Return $ret$\;
\end{small}
\caption{Exploration of number of memory sectors.}
\label{alg:sector_number}
\vspace*{2pt}
\end{algorithm}

Following the above-discussed procedures, we have generated \textit{15,233 configurations of the DESCNet architecture for the CapsNet}, and \textit{215,693 configurations for the DeepCaps}, with different design options (\textit{SMP}, \textit{SEP}, \textit{HY}), different sizes and number of sectors. 
Note that the \textit{SMP} and \textit{SEP} design options can also be considered as the boundary cases of the \textit{HY} design option. On one hand, a \textit{HY} organization where $SZ_D$, $SZ_W$ and $SZ_A$ are maximum is equivalent to the \textit{SEP}, because the corresponding $SZ_S$ for the \textit{HY} results to be null. On the other hand, if $SZ_D$, $SZ_W$ and $SZ_A$ of a \textit{HY} organization are all equal to 0, its resulting $SZ_S$ would have the same value as the one for the \textit{SMP}\footnote{Note that this particular solution, with $SZ_D = SZ_W = SZ_A = 0$, cannot be achieved for a HY solution, due to the minimum constraints given in Algorithm~\ref{alg:hybrid_size}. However, it represents a hypotetical extreme case to discuss.}.

\vspace*{-8pt}
\subsection{Our Methodology for the DSE of Scratchpad Memories}
\label{subsec:dse}

\begin{figure}[h]
	\centering
	\includegraphics[width=\linewidth]{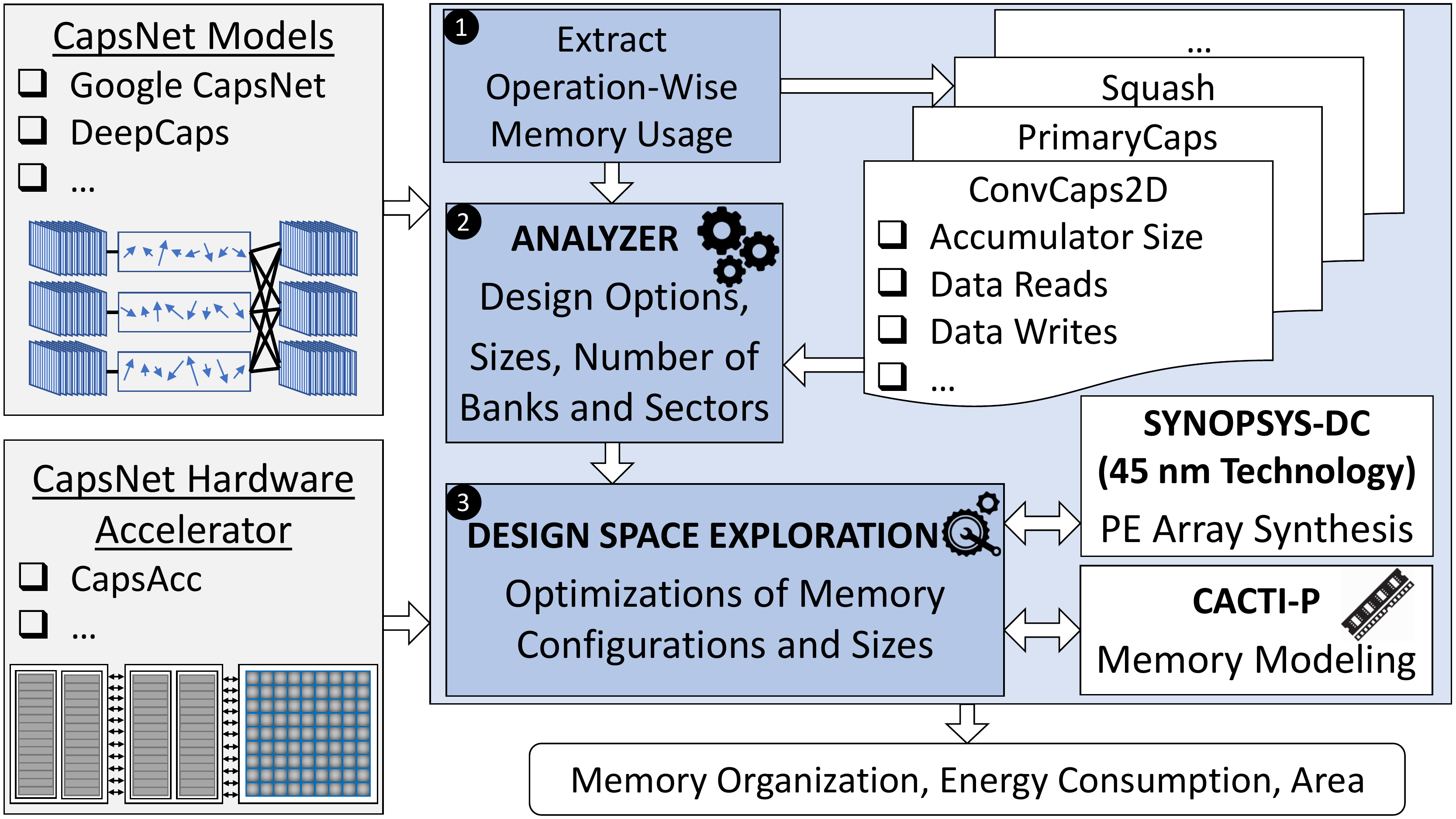}
	\caption{\textit{DESCNet} design space exploration and toolflow.}
	\label{fig:DESCNet_DSE_methodology}
	\vspace*{-3pt}
\end{figure}

The flow of our methodology is depicted in Fig.~\ref{fig:DESCNet_DSE_methodology}. 
Inputs are: CapsNet models and hardware accelerators for CapsNets. 
Output is: for each design option, the values of memory organization (i.e., size, number of banks and sectors), the energy consumption and area are generated. 
The key steps of our methodology are:

\begin{enumerate}[leftmargin=*]
    \item The extraction of the memory usage and memory accesses for each operation of the CapsNet inference. While the usage is needed for defining the design options and sizes, the read and write accesses, along with the operation-wise clock cycles, are used for computing the energy consumption.
    \item An analysis of the design options (\textit{SMP}, \textit{SEP}, \textit{HY}), and definition of the memory configurations, such as size and number of banks and sectors for the \mbox{power-gating}.
    \item A DSE of the possible memory configurations under analysis, through an exhaustive search, to find and select the non-dominated solutions. The estimation of area and energy consumption, with and without the \mbox{power-gating} option, are conducted through the CACTI-P tool~\cite{Li2011CACTI-P}. Note: we have performed an exhaustive search because, due to the practical limitations on the memory sizes and number of sectors, as discussed in Section~\ref{subsec:DSE_setup}, the execution time of the search still results relatively low\footnote{We measured the times for executing a complete DSE, including the estimation of energy and area provided by CACTI-P~\cite{Li2011CACTI-P}, of 1.5 minutes for the Google's CapsNet and of 22 minutes for the DeepCaps, when running with a snigle-thread application on an AMD Ryzen 5 CPU with 32GB RAM.}. However, if the search space increases, or more sophisticated memory evaluations require longer computational time, a heuristic search algorithm can easily be integrated into our methodology, in order to find a solution more quickly. Such a solution may be away from the optimal solution as found by the exhaustive search.
\end{enumerate}

\vspace*{-8pt}
\section{Evaluating our DESCNet Architectures}
\label{sec:results}

\subsection{Results for Google's CapsNets~\cite{Sabour2017dynamic_routing}: Area and Energy of the \mbox{On-Chip} Memory}
\label{subsec:results_onchip}

\begin{figure}[h]
	\centering
	\vspace*{3pt}
	\includegraphics[width=\linewidth]{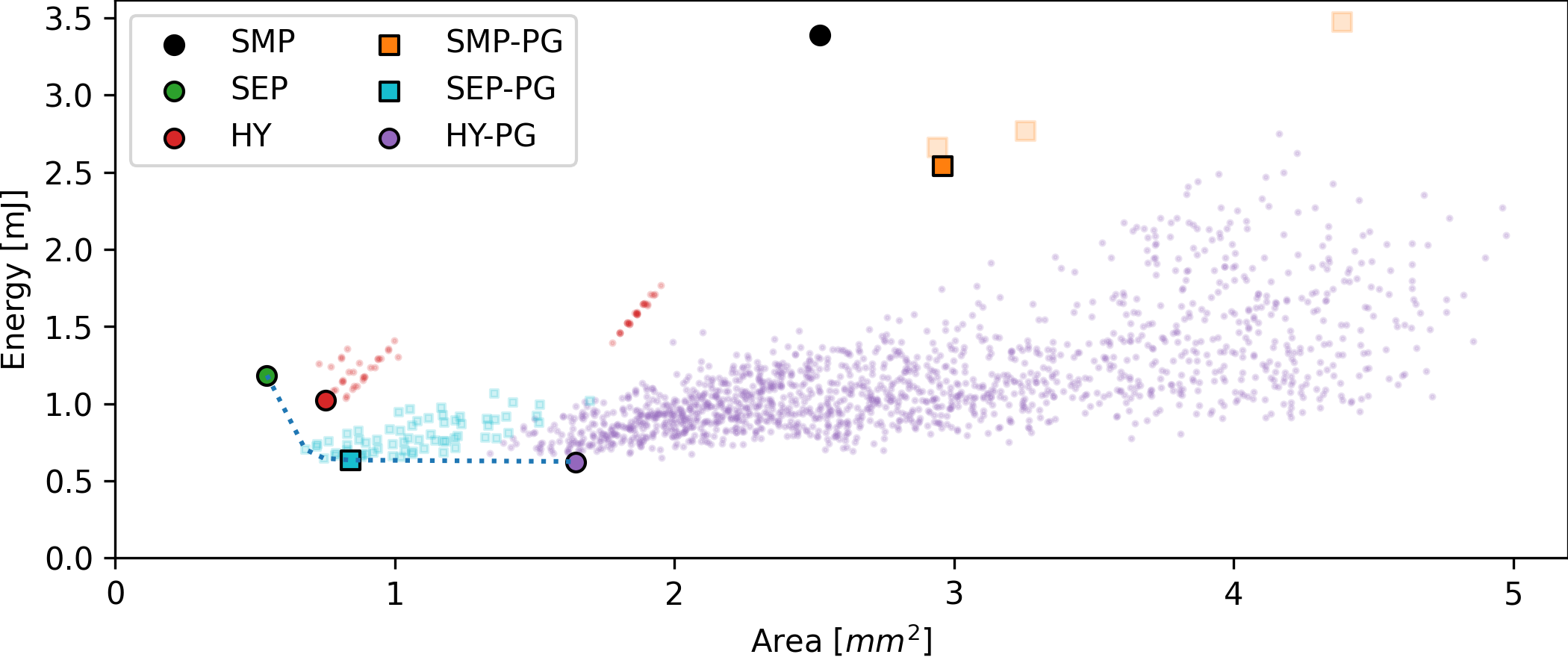}
	\vspace*{-14pt}
	\caption{DSE results of the \textit{DESCNet} memory configurations for the CapsNet.}
	\label{fig:dse_caps}
\end{figure}

We evaluate different memory architectural options (as discussed in~Section~\ref{sec:design}) for area and energy consumption, using the CACTI-P~\cite{Li2011CACTI-P} tool. The results of different {\it DESCNet} architectural designs of the scratchpad memory for the CapsNet on the MNIST dataset are discussed below. 
\vspace*{5pt}
\textbf{Design Space Exploration Results and Selected Configurations (Fig.~\ref{fig:dse_caps}):} 
The figure shows the trade-off between energy and area for 15,233 different \textit{DESCNet} architectural configurations. For each design option (\textit{SMP, SEP, HY}) and its corresponding version with \mbox{power-gating} (with suffix \textit{-PG}), the Pareto-optimal solutions with lowest-energy are selected. Note, while \textit{SEP}, \textit{SEP-PG} and \textit{HY-PG} belong to the Pareto-frontier, \textit{HY}, \textit{SMP} and \textit{SMP-PG} are dominated by other configurations. 
Their size and number of sectors are reported in Table~\ref{tab:configurations_caps}.

\begin{table}[h]
\caption{Selected memory configurations for the CapsNet.}
\label{tab:configurations_caps}
\resizebox{\columnwidth}{!}{
\begin{tabular}{|l|cc|cc|cc|cc|}
\toprule
\textbf{Mem} & \multicolumn{2}{c|}{\bf Shared} & \multicolumn{2}{c|}{\bf Data} & \multicolumn{2}{c|}{\bf Weight} & \multicolumn{2}{c|}{\bf Acc} \\
  &  SZ &  SC &  SZ &  SC &  SZ &  SC &  SZ &  SC \\
\midrule
    SEP &      --- &       --- &    25 kiB &          1 &      64 kiB &            1 &   32 kiB &         1 \\
 SEP-PG &      --- &       --- &    25 kiB &          2 &      64 kiB &            8 &   32 kiB &         2 \\
    SMP &  108 kiB &         1 &       --- &        --- &         --- &          --- &      --- &       --- \\
 SMP-PG &  108 kiB &         2 &       --- &        --- &         --- &          --- &      --- &       --- \\
     HY &   25 kiB &         1 &     8 kiB &          1 &      32 kiB &            1 &   16 kiB &         1 \\
  HY-PG &   32 kiB &         2 &    25 kiB &          2 &      25 kiB &            4 &   32 kiB &         2 \\
\bottomrule
\end{tabular}
}
\vspace*{3pt}
\end{table}

\begin{figure*}[t]
\centering
\includegraphics[width=\textwidth]{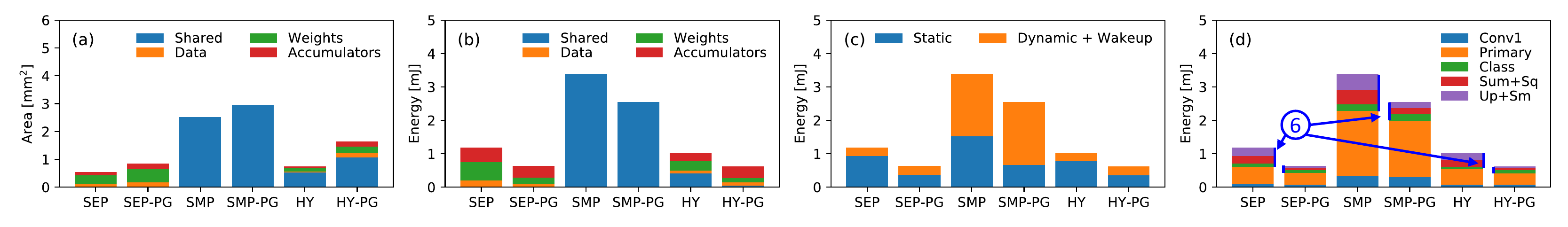}
\vspace*{-13pt}
\caption{Google's CapsNet Results for different components of the \textit{DESCNet} memory configurations: \textbf{(a)} Area breakdown, \textbf{(b)} Energy breakdown, \textbf{(c)} Static vs. dynamic energy consumption, \textbf{(d)} Energy breakdown of the different operations of the CapsNet inference.}
\label{fig:capsnet_4results}
\vspace*{0pt}
\end{figure*}

\textbf{Area Comparison (Fig.~\ref{fig:capsnet_4results}a):} 
The figure shows the area breakdown of different memory components of the {\it DESCNet}. 
We notice that, while the organization {\it SEP} has relatively larger memory sizes, compared to the other architectures, their area is relatively smaller. 
This effect is due to having single-port memories instead of a shared multi-ported design, where the latter requires more area for the complex interconnections. Indeed the area of the \textit{HY} organization is lower than the \textit{SMP}, due to a \mbox{small-sized} shared memory. 
Moreover, the \mbox{power-gating} circuitry also incurs additional area overhead (on average, 2.75\% for equally-sized SPMs) due to the sleep transistors. 


\vspace*{5pt}
\textbf{Energy Breakdown at the Component Level (Fig.~\ref{fig:capsnet_4results}b):} 
The figure shows that the design option {\it HY-PG} is more energy efficient than the others, due to having high flexibility and thus the higher potential of \mbox{power-gating} an heterogeneous combination of sectors, as compared to other designs such as the \textit{SEP-PG}, whose energy consumption is slightly higher. 
Note, despite having a smaller size than the weight memory, the shared memory of the \textit{HY} organization consumes a higher energy, due to the more complex internal architecture of a multi-ported memory.


\vspace*{5pt}
\textbf{Dynamic vs. Static Energy Consumption (Fig.~\ref{fig:capsnet_4results}c):} 
When comparing different architectural designs, the figure illustrates that: (1) moving from {\it SMP} to {\it SEP} and then to \textit{HY}, the dynamic energy can be reduced progressively; 
(2) moving from {\it HY} to {\it HY-PG}, the static energy can be further reduced, due to the benefits of the \mbox{power-gating}, 
and (3) the dynamic energy remains unchanged between \textit{non-PG} and \textit{-PG} organizations. 

Besides this, we noticed that the wakeup latency overhead is negligible. 
Even though it is masked by preloading the necessary values, its value is very low (0.072$ns$) compared to the average computational time of an operation (614$\mu s$) where the sleep transistors driving their corresponding memory sectors are in a steady state, either \textit{ON} or \textit{OFF}. 
This behavior also explains why the contribution of the wakeup energy (on average 1.6$nJ$), which appears during the transitions between \textit{OFF} and \textit{ON}, is low. 

\vspace*{5pt}
\textbf{Energy Breakdown for the Different Operations of the CapsNet Inference (Fig.~\ref{fig:capsnet_4results}d):} 
Though the absolute values vary, the relative proportions of the energy consumption by different operations of the {\it CapsNet} inference remain approximately similar across different memory designs. 
The highest portion of energy comes from the {\it PrimaryCaps (Prim)} layer, since it has a high memory utilization and frequent access to it,   
requiring most of the available memory, and thereby providing limited \mbox{power-gating} potential. 
On the contrary, the energy consumed by the dynamic routing operations (i.e., \textit{Sum+Squash}, \textit{Update+Softmax}) is significantly lower for the \textit{-PG} organizations (see pointer \rpoint{6} in Fig.~\ref{fig:capsnet_4results}).

For the detailed memory breakdown for the selected Pareto-optimal solutions, we refer to Appendix~\ref{app:mem_breakdown}.1.

\vspace*{-8pt}
\subsection{Results for DeepCaps~\cite{Rajasegaran2019deepcaps}: Area and Energy of the \mbox{On-Chip} Memory}
\label{subsec:results_deepcaps}

\begin{figure}[h]
	\centering
	\includegraphics[width=\linewidth]{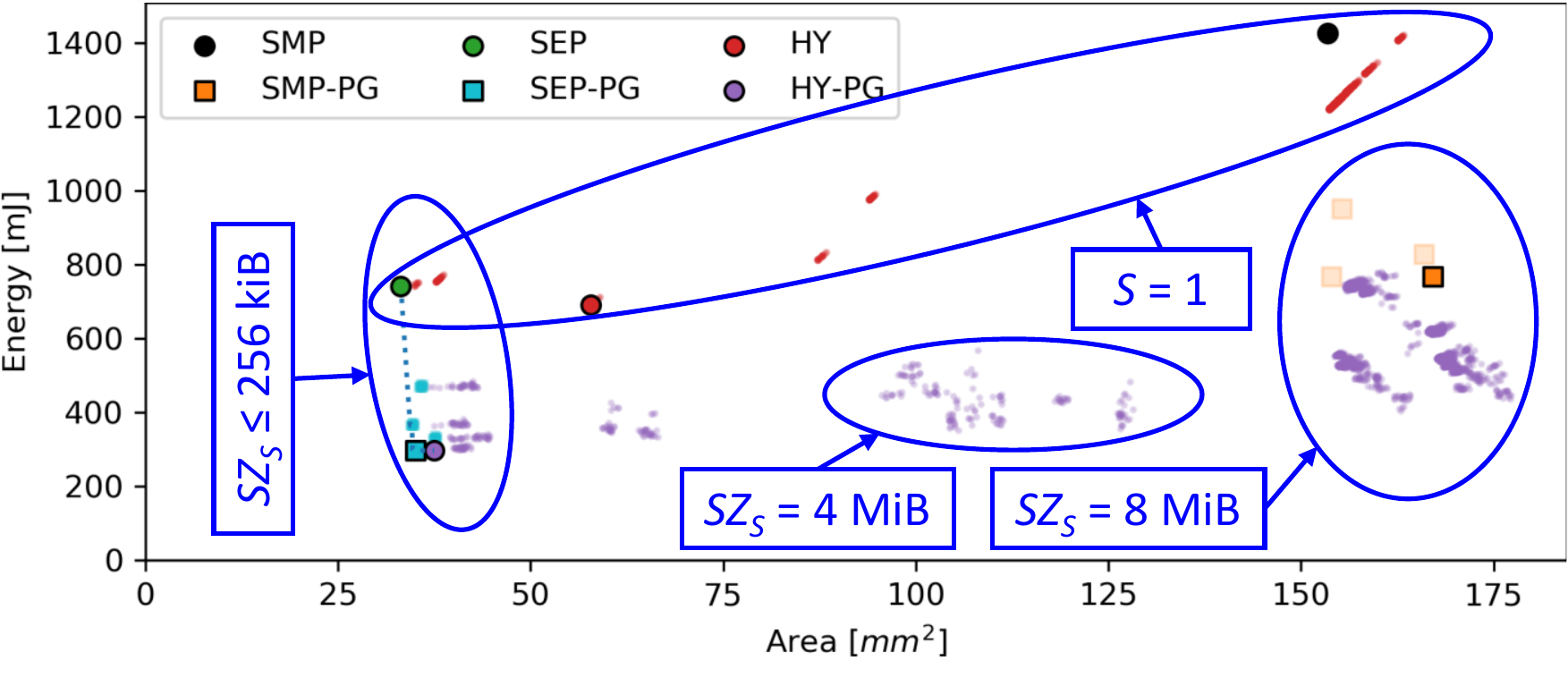}
	\caption{Design space exploration results of the \textit{DESCNet} memory configurations for the DeepCaps.}
	\label{fig:dse_deepcaps}
	\vspace*{-2pt}
\end{figure}

Similarly to the above-discussed results for the Google's CapsNet~\cite{Sabour2017dynamic_routing}, detailed evaluations are also conducted for the DeepCaps~\cite{Rajasegaran2019deepcaps}. Fig.~\ref{fig:dse_deepcaps} shows the solutions in the space area vs. energy, for 215,693 different memory design organizations. 
The selected solutions with memory size and number of sectors are reported in Table~\ref{tab:configurations_deepcaps}. Note that the high-energy solutions have only 1 sector (i.e., the power-gating cannot be applied), the high-area solutions have the size of the shared memory equal to 8 \textit{MiB}, while the solutions with low area and low energy consumption have a shared memory with a size lower than or equal to 256 \textit{kiB}. 
Compared to the CapsNet, the DeepCaps needs larger memory sizes, to efficiently handle the large-scale and more-complex computations. 

\begin{figure*}[h]
\centering
\includegraphics[width=\textwidth]{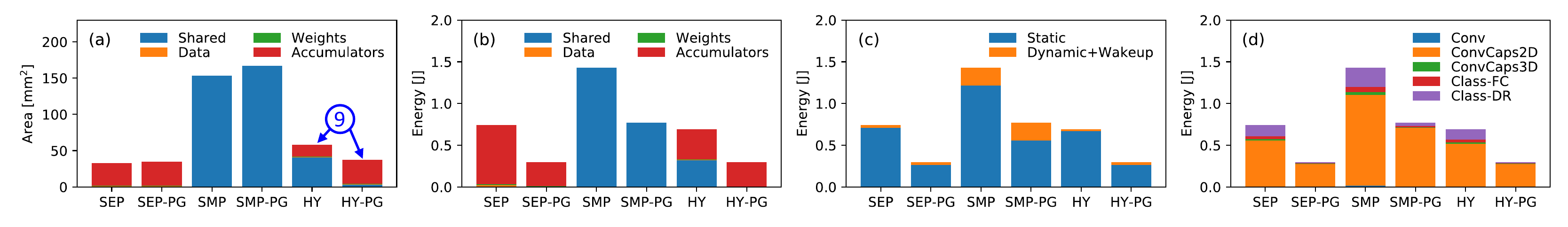}
\vspace*{-13pt}
\caption{DeepCaps results for different components of the \textit{DESCNet} memory congigurations: \textbf{(a)} Area breakdown, \textbf{(b)} Energy breakdown, \textbf{(c)} Static vs. dynamic energy consumption, \textbf{(d)} Energy breakdown of the different operations of the DeepCaps inference.}
\label{fig:deepcaps_4results}
\vspace*{-2pt}
\end{figure*}

\begin{table}[h]
\vspace*{2pt}
\caption{Selected memory configurations for the DeepCaps.}
\label{tab:configurations_deepcaps}
\resizebox{\columnwidth}{!}{
\begin{tabular}{|l|cc|cc|cc|cc|}
\toprule
\textbf{Mem} & \multicolumn{2}{c|}{\bf Shared} & \multicolumn{2}{c|}{\bf Data} & \multicolumn{2}{c|}{\bf Weight} & \multicolumn{2}{c|}{\bf Acc} \\
  &  SZ &  SC &  SZ &  SC &  SZ &  SC &  SZ &  SC \\
\midrule
    SEP &      --- &       --- &   256 kiB &          1 &     128 kiB &            1 &    8 MiB &         1 \\
 SEP-PG &      --- &       --- &   256 kiB &          8 &     128 kiB &           16 &    8 MiB &        16 \\
    SMP &    8 MiB &         1 &       --- &        --- &         --- &          --- &      --- &       --- \\
 SMP-PG &    8 MiB &         8 &       --- &        --- &         --- &          --- &      --- &       --- \\
     HY &    2 MiB &         1 &   108 kiB &          1 &       8 kiB &            1 &    4 MiB &         1 \\
  HY-PG &  128 kiB &         2 &   128 kiB &          8 &      64 kiB &            8 &    8 MiB &        16 \\\hline
       HY, P\textsubscript{S}=1 &    4 MiB &         1 &   256 kiB &          1 &       8 kiB &            1 &    2 MiB &         1 \\
  HY-PG, P\textsubscript{S}=1 &    4 MiB &         8 &   256 kiB &          8 &     128 kiB &           16 &    2 MiB &         4 \\
\bottomrule
\end{tabular}
}
\vspace*{2pt}
\end{table}

As a consequence, although allowing to explore more different solutions, having larger memory sizes implies higher area and energy consumption; see results in Fig.~\ref{fig:deepcaps_4results}. The memory breakdown is discussed in detail in Appendix~\ref{app:mem_breakdown}.2. 
Regarding the area, an interesting result is the lower area of the \textit{HY-PG} compared to the \textit{HY} (see pointer \rpoint{9} in Fig.~\ref{fig:deepcaps_4results}), despite the \mbox{power-gating} circuitry. This is due to the different sizes of shared memory and accumulators, which are the most impactful memories, between the two organizations. 

The energy consumption for the \textit{-PG} organization is significantly lower than the \textit{non-PG} counterparts, due to a heterogeneous usage of the memories across different operations of the DeepCaps. Hence, applying \mbox{power-gating} reduces the static energy not only for the dynamic routing operations, but also for the ConvCaps2D computations, whose contribution is the highest, as shown in Fig.~\ref{fig:deepcaps_4results}d. 

In a similar way as noticed for the Google's CapsNet, for the DeepCaps, the \textit{HY-PG} is the solution with the lowest energy consumption, the \textit{SEP} organization has the lowest area, and the \textit{SEP-PG} is another organization belonging to the Pareto-frontier.

\vspace*{-6pt}
\subsection{DSE for the HY-PG Design Option with Size-Constrained Memory for DeepCaps~\cite{Rajasegaran2019deepcaps}}
\label{subsec:DSE_constraint}
\vspace*{-2pt}

Motivated by the observation that the shared memory size has a great impact on the efficiency (see Fig.~\ref{fig:dse_deepcaps}), and to the fact that embedded systems might have size-constrained memory, we extend the analysis by exploring the HY-PG architectural organizations with a memory constraint. More specifically, we performed a DSE by constraining the maximum size of the shared memory. 
Moreover, the memory usage patterns and partitions reported in Appendix~\ref{app:mem_breakdown}.2 show that the shared memory of the \textit{HY} and \textit{HY-PG} design options do not always require having three ports, because for some solutions, the shared memory only needs to store one or two different types of values.

\begin{figure}[h]
	\centering
	\includegraphics[width=\linewidth]{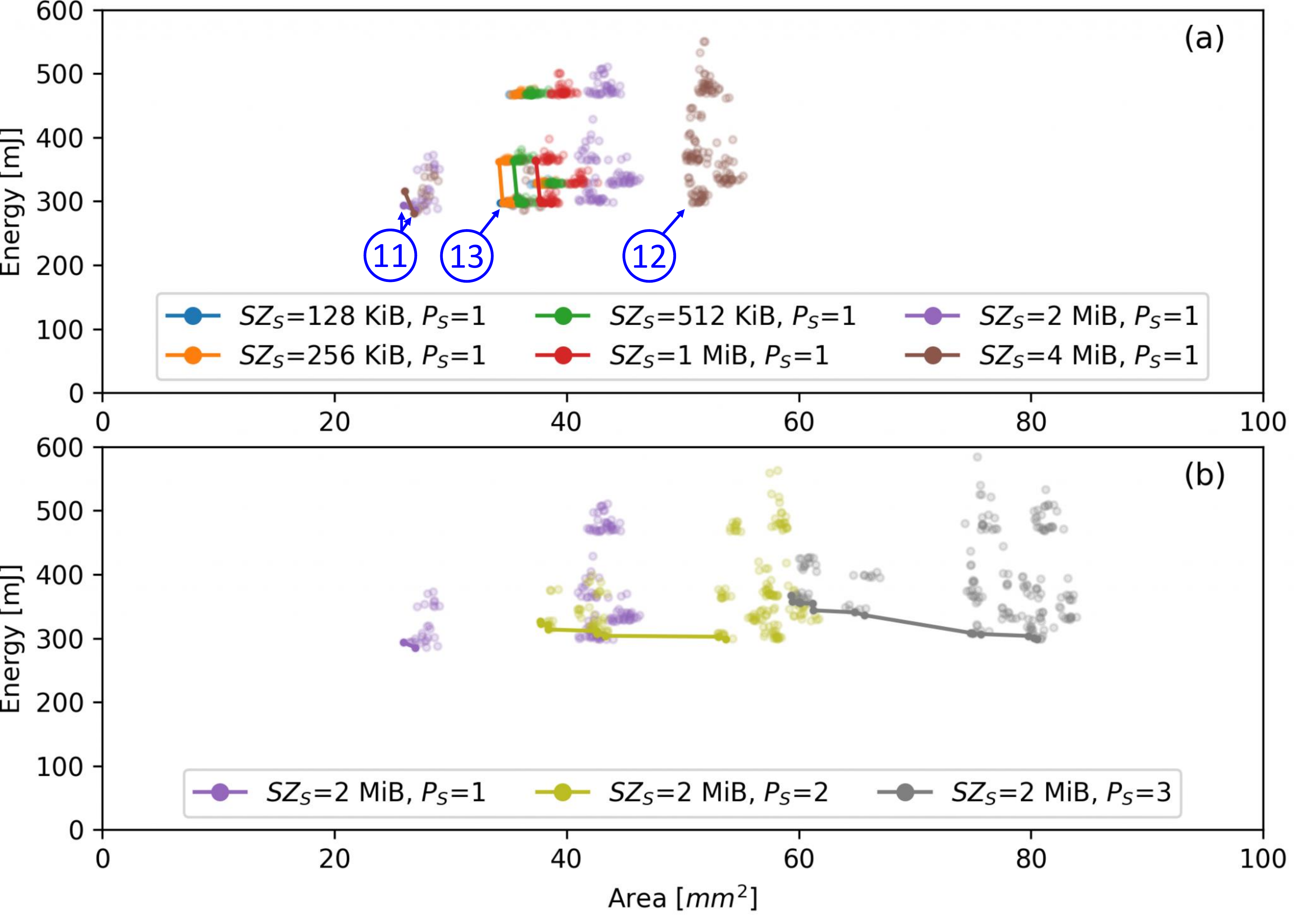}
	\caption{Design space exploration results for the \textit{HY-PG DESCNet} memory configurations for the DeepCaps, having constraints on the size and number of ports of the shared memory.}
	\label{fig:dse_deepcaps_dc}
	\vspace*{-3pt}
\end{figure}

To this regard, in this analysis we also explored the space of the \textit{HY-PG} solutions when the number of ports of the shared memory ($P_S$) is constrained. Fig.~\ref{fig:dse_deepcaps_dc} shows their tradeoffs between area and energy consumption, for 113,337 different memory configurations. The most efficient solutions have a size of the 1-port shared memory equal to $2~MiB$ and $4~MiB$ (see pointer \rpoint{11} in Fig.~\ref{fig:dse_deepcaps_dc}a), while the worst results are obtained by the combination of a shared memory of $4~MiB$ and accumulator memory of $8~MiB$ (see pointer \rpoint{12}). Note, despite having a smaller-sized shared memory, the solutions with the 1-port shared memory of size $128~kiB$ and $256~kiB$ (see pointer \rpoint{13}) are relatively less efficient. This implies that such a size of the shared memory can more efficiently couple with the rest of the system, to achieve overall a lower area and energy consumption. Moreover, as clearly visible from Figure~\ref{fig:dse_deepcaps_dc}b, the area and energy efficiency is improved by having a lower $P_S$. The detailed memory configuration for the \textit{HY} and \textit{HY-PG} lowest-energy solutions are shown in the respective lines of Table~\ref{tab:configurations_deepcaps}.

\begin{table*}[h]
\centering
\caption{Area and energy consumption results for different \textit{DESCNet} architectural organizations.}
\label{tab:results_final}
\resizebox{\linewidth}{!}{%
\begin{tabular}{|cc|c|c|c|c|c|c|c|c|c|c|c|c|c|c|c|c|}
\hline
\multirow{4}{*}{\bf NN} & \multirow{4}{*}{\bf Mem} & \multicolumn{4}{c|}{\textbf{Shared Mem}} & \multicolumn{4}{c|}{\textbf{Weight Mem}} & \multicolumn{4}{c|}{\textbf{Data Mem}} & \multicolumn{4}{c|}{\textbf{Accumulator Mem}} \\
& &  \multicolumn{1}{c}{} & \multicolumn{1}{c}{\textbf{Dynamic}} & \multicolumn{1}{c}{\textbf{Static}} & \multicolumn{1}{c|}{\textbf{Wakeup}}& \multicolumn{1}{c}{} & \multicolumn{1}{c}{\textbf{Dynamic}} & \multicolumn{1}{c}{\textbf{Static}} & \multicolumn{1}{c|}{\textbf{Wakeup}}& \multicolumn{1}{c}{} & \multicolumn{1}{c}{\textbf{Dynamic}} & \multicolumn{1}{c}{\textbf{Static}} & \multicolumn{1}{c|}{\textbf{Wakeup}}& \multicolumn{1}{c}{} & \multicolumn{1}{c}{\textbf{Dynamic}} & \multicolumn{1}{c}{\textbf{Static}} & \multicolumn{1}{c|}{\textbf{Wakeup}} \\
 & & \multicolumn{1}{c}{\textbf{Area}} & \multicolumn{1}{c}{\textbf{Energy}} & \multicolumn{1}{c}{\textbf{Energy}} & \multicolumn{1}{c|}{\textbf{Energy}}& \multicolumn{1}{c}{\textbf{Area}} & \multicolumn{1}{c}{\textbf{Energy}} & \multicolumn{1}{c}{\textbf{Energy}} & \multicolumn{1}{c|}{\textbf{Energy}}& \multicolumn{1}{c}{\textbf{Area}} & \multicolumn{1}{c}{\textbf{Energy}} & \multicolumn{1}{c}{\textbf{Energy}} & \multicolumn{1}{c|}{\textbf{Energy}}& \multicolumn{1}{c}{\textbf{Area}} & \multicolumn{1}{c}{\textbf{Energy}} & \multicolumn{1}{c}{\textbf{Energy}} & \multicolumn{1}{c|}{\textbf{Energy}} \\
 & & \multicolumn{1}{c}{\textbf{{[}mm2{]}}} & \multicolumn{1}{c}{\textbf{{[}mJ{]}}} & \multicolumn{1}{c}{\textbf{{[}mJ{]}}} & \multicolumn{1}{c|}{\textbf{{[}nJ{]}}}& \multicolumn{1}{c}{\textbf{{[}mm2{]}}} & \multicolumn{1}{c}{\textbf{{[}mJ{]}}} & \multicolumn{1}{c}{\textbf{{[}mJ{]}}} & \multicolumn{1}{c|}{\textbf{{[}nJ{]}}}& \multicolumn{1}{c}{\textbf{{[}mm2{]}}} & \multicolumn{1}{c}{\textbf{{[}mJ{]}}} & \multicolumn{1}{c}{\textbf{{[}mJ{]}}} & \multicolumn{1}{c|}{\textbf{{[}nJ{]}}}& \multicolumn{1}{c}{\textbf{{[}mm2{]}}} & \multicolumn{1}{c}{\textbf{{[}mJ{]}}} & \multicolumn{1}{c}{\textbf{{[}mJ{]}}} & \multicolumn{1}{c|}{\textbf{{[}nJ{]}}} \\ \hline
\parbox[t]{2mm}{\multirow{6}{*}{\rotatebox[origin=c]{90}{CapsNet}}} &     SEP &         --- &                   --- &                  --- &              --- &       \textbf{0.314} &                 0.051 &                0.501 &              --- &     \textbf{0.104} &               0.011 &              0.188 &            --- &    \textbf{0.125} &              0.196 &             0.238 &           --- \\
  &  SEP-PG &         --- &                   --- &                  --- &              --- &       0.469 &                 0.053 &                0.135 &            0.044 &     0.173 &               0.012 &              0.083 &          0.048 &    0.200 &              0.205 &             0.148 &         0.064 \\
  &     SMP &       2.521 &                 1.859 &                1.529 &              --- &         --- &                   --- &                  --- &              --- &       --- &                 --- &                --- &            --- &      --- &                --- &               --- &           --- \\
  &  SMP-PG &       2.958 &                 1.875 &                0.668 &            0.352 &         --- &                   --- &                  --- &              --- &       --- &                 --- &                --- &            --- &      --- &                --- &               --- &           --- \\
  &      HY &       0.519 &                 0.068 &                0.348 &              --- &       0.125 &                 0.044 &                0.238 &              --- &     0.041 &               0.009 &              0.068 &            --- &    0.067 &              0.120 &             0.130 &           --- \\
  &   HY-PG &       1.061 &                \textbf{ 0.004} &                \textbf{0.046} &            0.080 &       0.213 &                 \textbf{0.045} &                \textbf{0.083} &            0.032 &     0.173 &              \textbf{ 0.012} &              \textbf{0.083} &          0.048 &    0.200 &              \textbf{0.205} &            \textbf{ 0.148} &         0.064 \\\hline
  
\parbox[t]{2mm}{\multirow{8}{*}{\rotatebox[origin=c]{90}{DeepCaps}}} &     SEP &         --- &                   --- &                  --- &              --- &      0.617 &                 0.039 &               12.172 &              --- &     1.165 &               0.098 &             22.266 &            --- &   31.392 &             34.268 &           673.562 &           --- \\
 &  SEP-PG &         --- &                   --- &                  --- &              --- &       0.896 &                 0.040 &                2.277 &            0.044 &     1.223 &               0.099 &              4.695 &          0.247 &   32.905 &             34.464 &           256.029 &         0.642 \\
 &     SMP &     153.474 &               213.961 &             1214.223 &              --- &         --- &                   --- &                  --- &              --- &       --- &                 --- &                --- &            --- &      --- &                --- &               --- &           --- \\
 &  SMP-PG &     167.077 &               214.115 &              555.102 &            1.234 &         --- &                   --- &                  --- &              --- &       --- &                 --- &                --- &            --- &      --- &                --- &               --- &           --- \\
  &      HY &      41.067 &                 3.143 &              315.671 &              --- &       0.041 &                 0.003 &                0.810 &              --- &     0.547 &               0.074 &             10.453 &            --- &   16.168 &             20.954 &           339.713 &           --- \\
  &   HY-PG &       3.293 &                 0.125 &                1.211 &            0.464 &       0.469 &                 0.019 &                1.619 &            0.044 &     0.816 &               0.079 &              3.898 &          0.064 &   32.905 &             34.464 &           256.029 &         0.642 \\
  
    &      HY, P\textsubscript{S}=1 &      16.168 &                 6.108 &              339.713 &              --- &       0.041 &                 0.003 &                0.810 &              --- &     1.165 &               0.098 &             22.266 &            --- &    8.949 &             11.037 &           173.698 &           --- \\
    &   HY-PG, P\textsubscript{S}=1 &      \textbf{17.731} &                 \textbf{6.019} &              \textbf{120.913} &            0.642 &      \textbf{ 0.896} &                 \textbf{0.040} &                \textbf{2.277} &            0.044 &     \textbf{1.223} &              \textbf{ 0.099} &              \textbf{4.695} &          0.247 &    \textbf{8.338} &             \textbf{11.091} &           \textbf{128.696} &         0.642 \\
\hline
\end{tabular}%
}
\vspace*{6pt}
\end{table*}

\vspace*{-6pt}
\subsection{Impact of the Memory on the Complete CapsNet Accelerator Architecture}
\label{subsec:results_complete}
\vspace*{-2pt}

Based on the evaluations performed in Section~\ref{subsec:results_onchip}, we select two Pareto-optimal {\it DESCNet} architectures for the CapsNet, which are {\it SEP} and \textit{HY-PG}. 
The choice of these organizations is strategical, because they represent the Pareto-optimal solutions with the lowest area and the lowest energy, respectively. 
Note that there is no performance loss, compared to the CapsNet and DeepCaps executed on the baseline \textit{CapsAcc}~\cite{Marchisio2019CapsAcc}. 
We synthesize the complete architecture of the DNN accelerator executing CapsNets for the MNIST dataset and DeepCaps on the CIFAR10 dataset in a 32nm CMOS technology library, using the ASIC design flow with the Synopsys Design Complier. 

\begin{figure}[h]
\centering
\vspace*{2pt}
\begin{minipage}[t]{.48\linewidth}
\vspace*{0mm}
\subfloat[]{
\includegraphics[width=\linewidth]{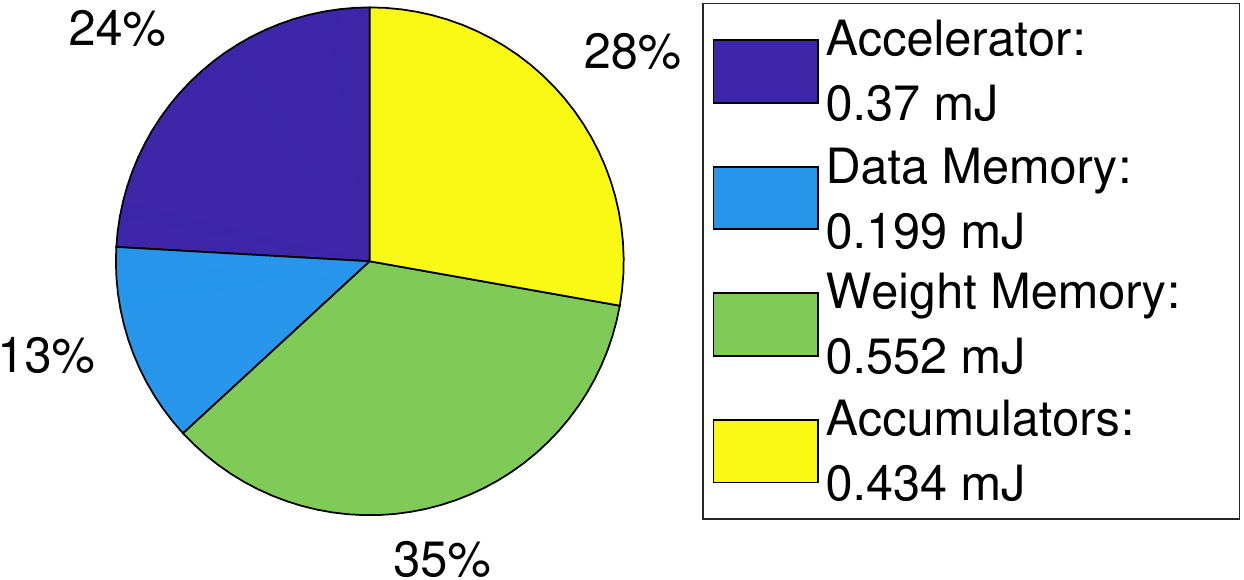}
\label{fig:energy_breakdown_SEP}}
\end{minipage}
\hfill
\begin{minipage}[t]{.48\linewidth}
\vspace*{0pt}
\subfloat[]{
\includegraphics[width=\linewidth]{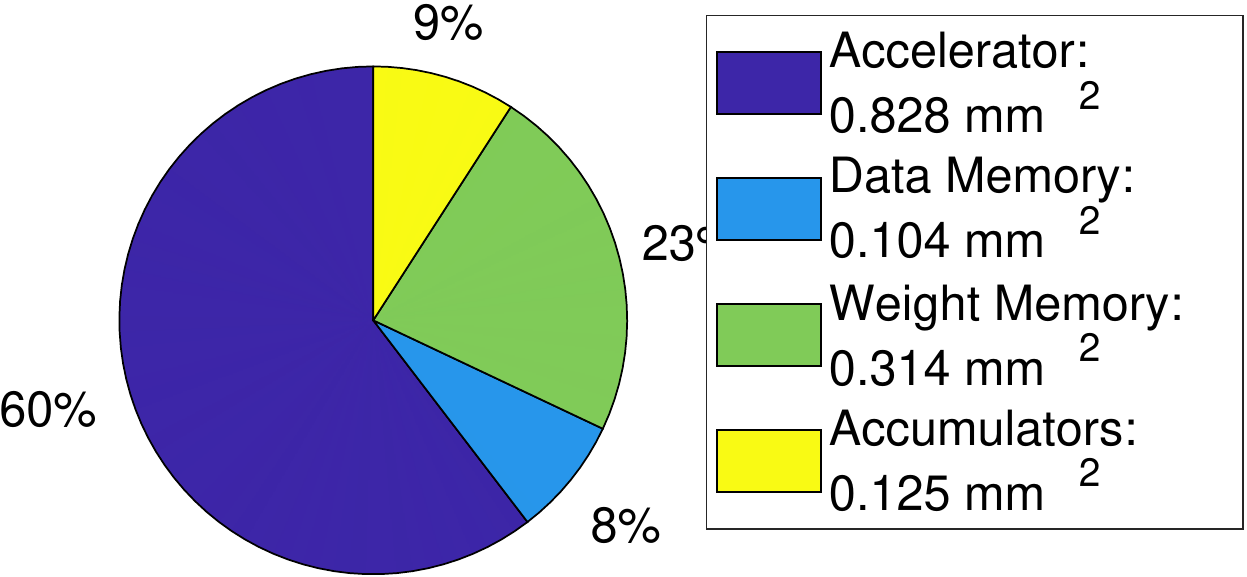}
\label{fig:area_breakdown_SEP}}
\end{minipage}
\caption{\textbf{(a)} Energy and \textbf{(b)} area breakdown of our CapsNet inference architecture using \textit{\textbf{SEP}} memory.}
\label{fig:energy_area_breakdown_SEP}
\end{figure}

\begin{figure}[h]
\vspace*{-3pt}
\centering
\begin{minipage}[t]{.48\linewidth}
\subfloat[]{
\includegraphics[width=\linewidth]{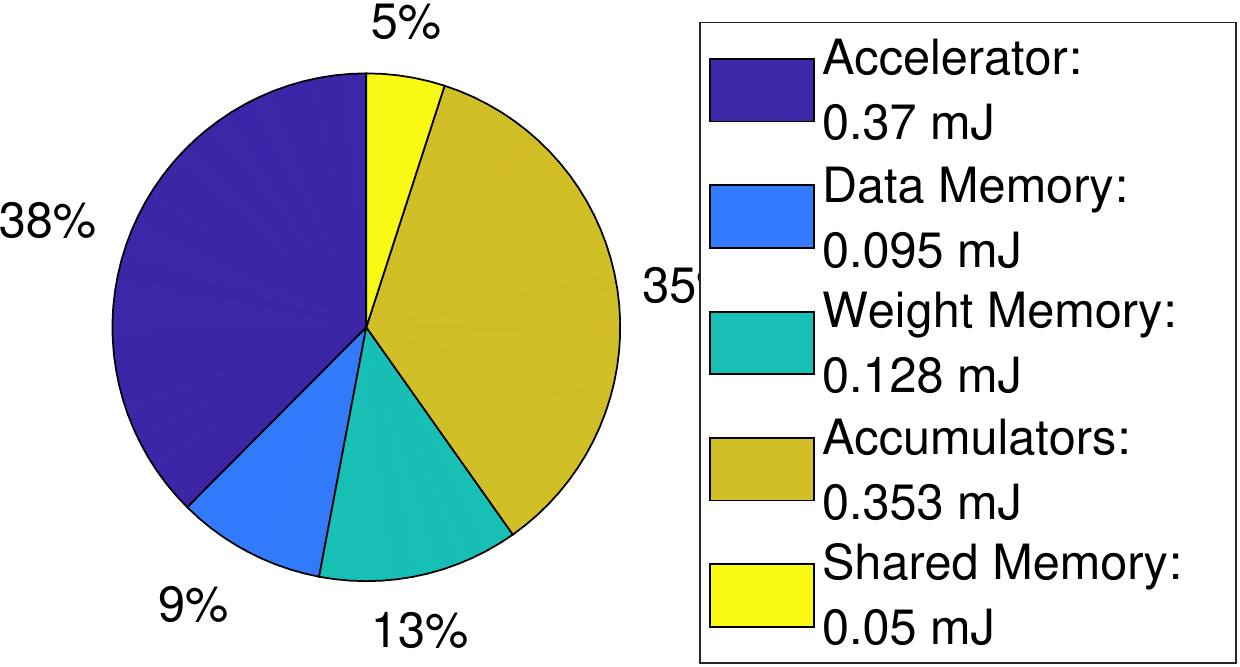}
\label{fig:energy_breakdown_HYPG}}
\end{minipage}
\hfill
\begin{minipage}[t]{.48\linewidth}
\subfloat[]{
\includegraphics[width=\linewidth]{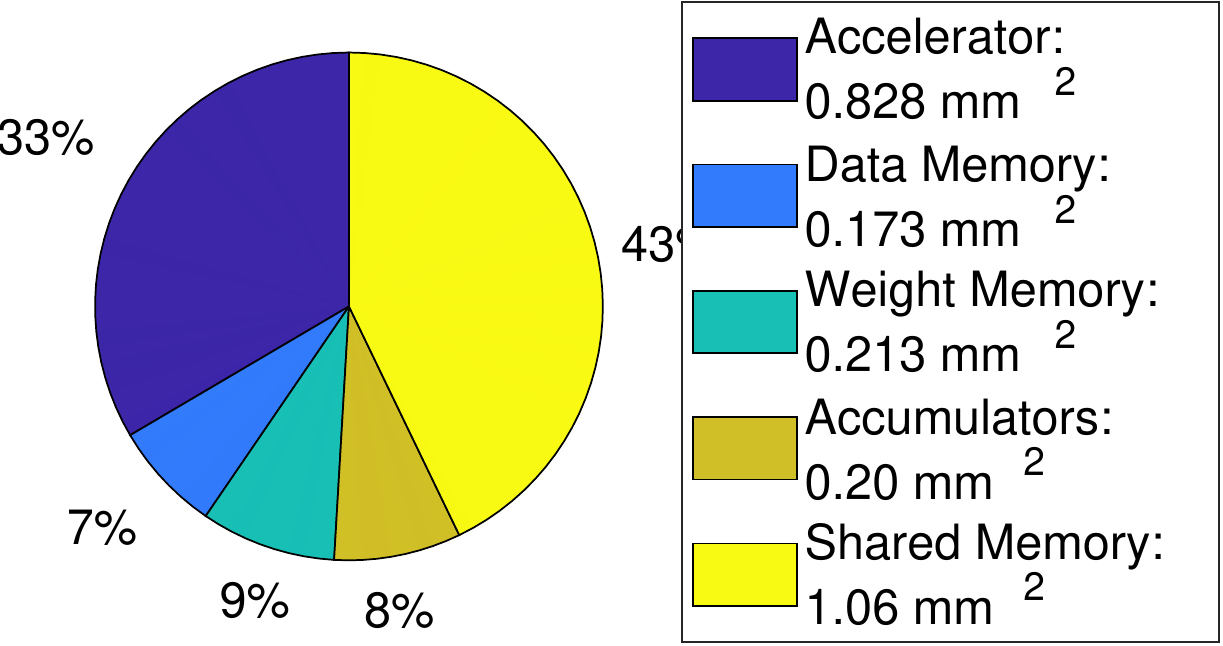}
\label{fig:area_breakdown_HYPG}}
\end{minipage}
\caption{\textbf{(a)} Energy and \textbf{(b)} area breakdown of our CapsNet inference architecture using \textit{\textbf{HY-PG}} memory.}
\label{fig:energy_area_breakdown_HYPG}
\vspace*{-5pt}
\end{figure}

The detailed area and energy estimations of the complete \mbox{on-chip} architectures for the Google's CapsNet, comprising the accelerator and the \mbox{on-chip} memories with \textit{SEP-PG} and \textit{HY-PG} organizations, are shown in Figures~\ref{fig:energy_area_breakdown_SEP} and~\ref{fig:energy_area_breakdown_HYPG}, respectively.

When comparing the initial design as discussed in Section~\ref{subsec:energy_breakdown} version (b) with the \textit{SEP} (Fig.~\ref{fig:energy_area_breakdown_SEP}), the total \mbox{on-chip} memory energy is reduced by $65\%$ and the \mbox{on-chip} memory area by $91\%$. 
Compared to version (a), our {\it SEP DESCNet} incurs $78\%$ reduced energy and $47\%$ reduced area for the complete accelerator. 

Our proposed \textit{DESCNet HY-PG} organization reduces the \mbox{on-chip} energy by $82\%$ and the \mbox{on-chip} area by $35\%$, compared to the version (b), while reducing the total energy and the total area by $79\%$ and $40\%$ compared to the version (a), respectively.

Consequently, compared to version (a), which corresponds to the state-of-the-art design of~\cite{Marchisio2019CapsAcc}, our proposed approach can provide total energy and area reductions, comprising accelerator, \mbox{on-chip} memory and \mbox{off-chip} memory, of up to $79\%$ and $47\%$, respectively, without any performance loss.

\begin{figure}[h]
\centering
\begin{minipage}[t]{.46\linewidth}
\subfloat[]{
\includegraphics[width=\linewidth]{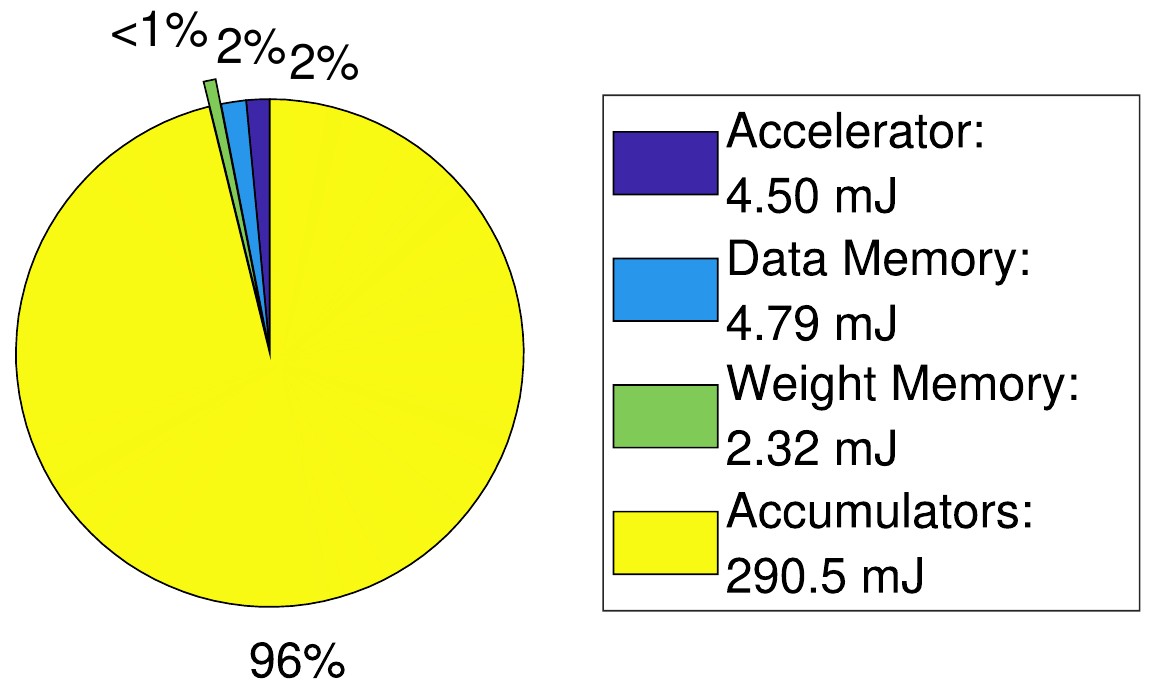}
\label{fig:energy_breakdown_SEPPG}}
\end{minipage}
\hfill
\begin{minipage}[t]{.46\linewidth}
\subfloat[]{
\includegraphics[width=\linewidth]{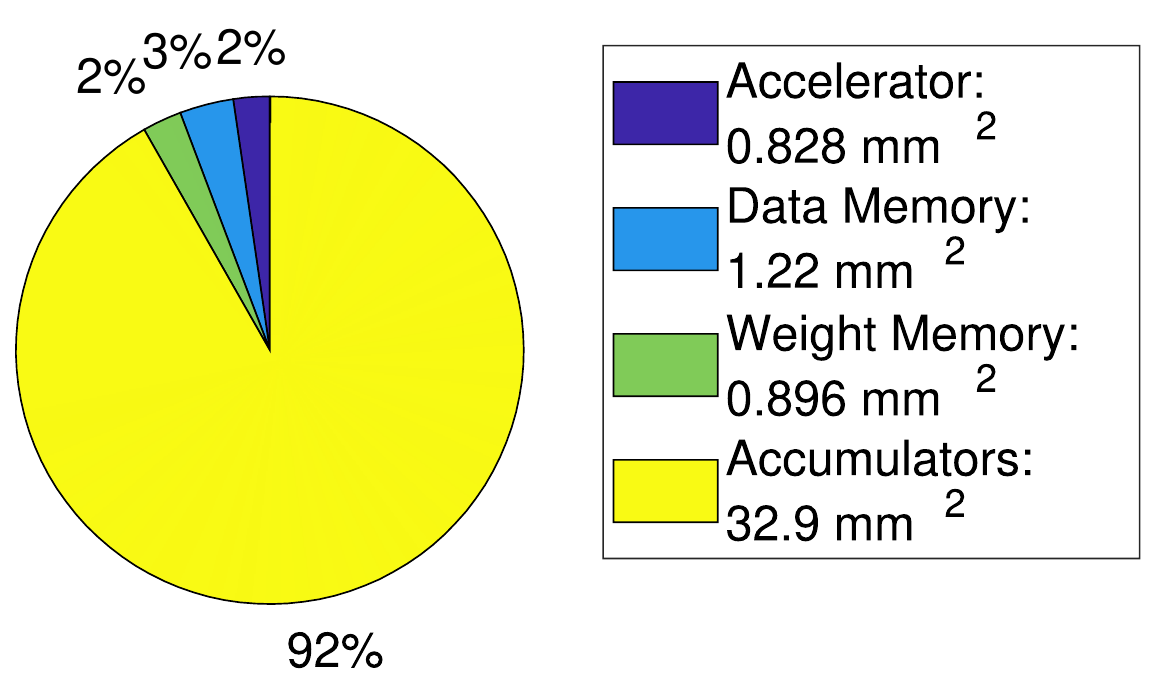}
\label{fig:area_breakdown_SEPPG}}
\end{minipage}
\caption{\textbf{(a)} Energy and \textbf{(b)} area breakdown of our DeepCaps inference architecture using \textit{\textbf{SEP-PG}} memory.}
\label{fig:energy_area_breakdown_SEPPG}
\vspace*{-4pt}
\end{figure}
\begin{figure}[h]
\centering
\begin{minipage}[t]{.48\linewidth}
\subfloat[]{
\includegraphics[width=\linewidth]{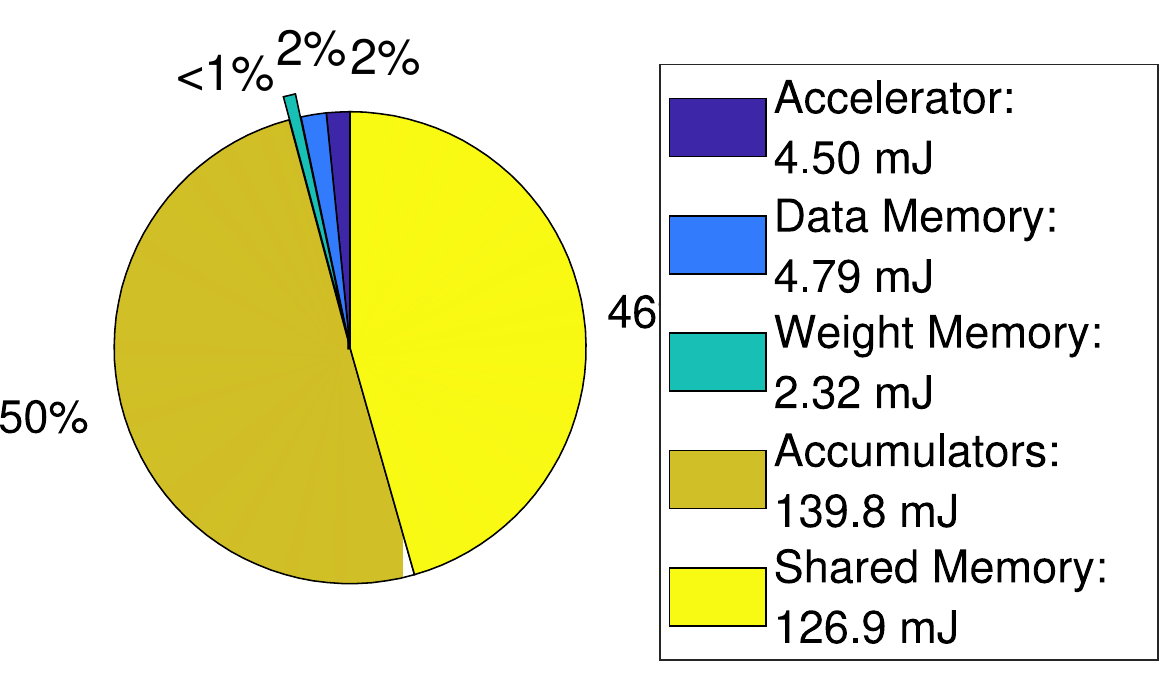}
\label{fig:energy_breakdown_HYPG_PS1}}
\end{minipage}
\hfill
\begin{minipage}[t]{.48\linewidth}
\subfloat[]{
\includegraphics[width=\linewidth]{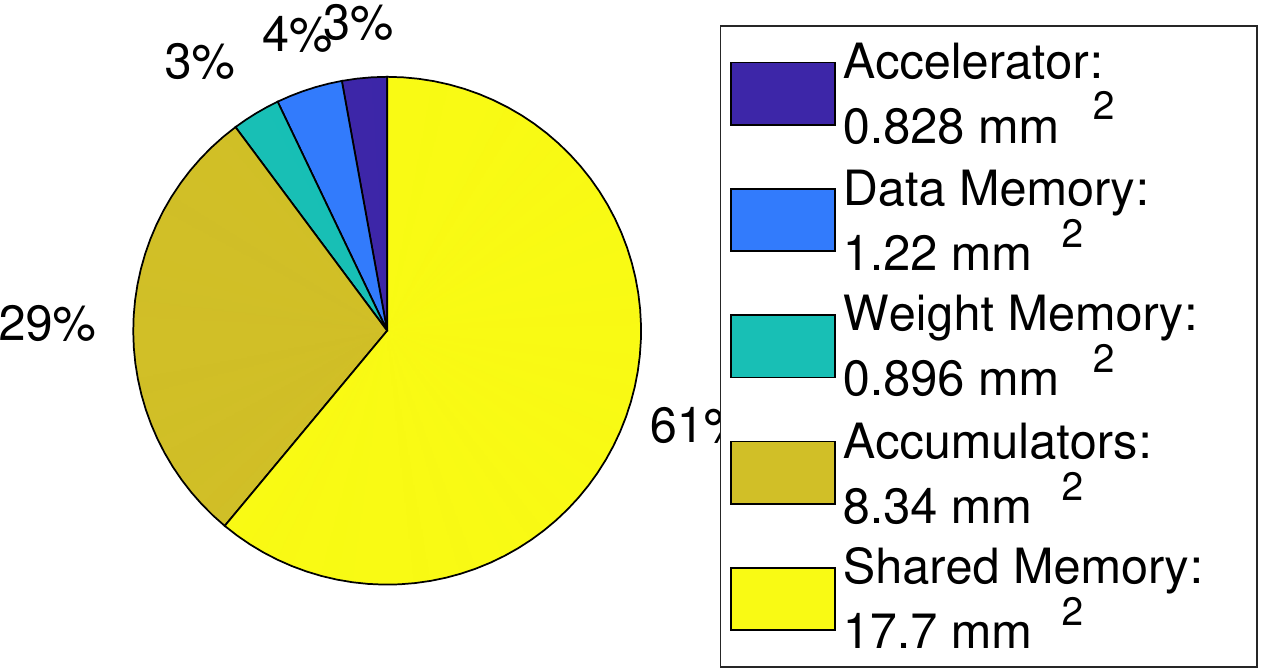}
\label{fig:area_breakdown_HYPG_PS1}}
\end{minipage}
\caption{\textbf{(a)} Energy and \textbf{(b)} area breakdown of our DeepCaps inference architecture using \textit{\textbf{HY-PG, P\textsubscript{S}=1}} memory.}
\label{fig:energy_area_breakdown_HYPG_PS1}
\vspace*{0pt}
\end{figure}

Energy and area estimations of the complete \mbox{on-chip} architecture executing the DeepCaps, using the \textit{SEP-PG} organization, are reported in Fig.~\ref{fig:energy_area_breakdown_SEPPG}. The graph shows a clear dominance of the accumulator memory on both the \mbox{on-chip} area and the \mbox{on-chip} energy, due to its large size. 
On the contrary, the \textit{HY-PG, P\textsubscript{S}=1} organization results reported in Fig.~\ref{fig:energy_area_breakdown_HYPG_PS1} show a relatively more balanced distribution of the energy and area between the accumulator and the shared memory. 
As previously discussed in Section~\ref{subsec:energy_breakdown}, since the original baseline architecture of~\cite{Marchisio2019CapsAcc} cannot execute the DeepCaps for the CIFAR10 dataset, we cannot compare their work with our DESCNet design deploying the DeepCaps. 
This also shows that our memory sub-system can indeed successfully support DeepCaps enabling its memory-efficient acceleration, which is not possible for the original baseline CapsAcc~\cite{Marchisio2019CapsAcc}.

\vspace*{-8pt}
\subsection{Results Summary and Discussion}
\label{subsec:result discussion}

Table~\ref{tab:results_final} shows the detailed results of the area and energy consumption for the different \textit{DESCNet} architectures, obtained by our DSE, for the CapsNet and the DeepCaps. 
The following \textbf{key observations} can be derived from our analyses.

\begin{itemize}[leftmargin=*]
    \item Despite many efforts in optimizing the computational arrays of the DNN accelerators, more fruitful area and energy savings are obtained when the optimizations are applied to the memory, as showcased in this paper with the DNN accelerators executing both the Google's CapsNets and the DeepCaps.
    \item Having large and shared multi-port memories, where different values are stored, e.g., the \textit{SMP} organization, is in general a bad design choice, because of the resource-hungry hardware overhead for handling heterogeneous accesses, which are not necessary if we systematically study the application-driven memory resource requirements. Indeed, as highlighted in Fig.~\ref{fig:dse_deepcaps}, having $SZ_S \leq 256\ kiB$ can relatively reduce jointly the energy and the area, compared to other solutions with larger $SZ_S$. 
    \item For a certain set of solutions belonging to the \textit{HY} and \textit{HY-PG} design options, the multi-port shared memory can be replaced by an equivalent single-port, offering further energy and area reductions due to a more balanced memory breakdown between the accumulator and the shared memory. 
    \item Employing efficient \mbox{on-chip} SPM memory organization architectures (e.g., \textit{SEP}, \textit{SEP-PG}, and \textit{HY-PG}) significantly reduces the hardware resource requirements, thereby bearing the development of DNN accelerator executing complex operations such as the Capsule Networks in resouce-constrained scenarios, which are typical for IoT-edge devices.
\end{itemize}

\vspace*{-3pt}
\section{Conclusion}
\label{sec:conclusions}

Based on an extensive analysis of the {\it CapsNet} inference, we identified that a significant amount of energy can be saved by designing a specialized memory hierarchy. 
To achieve high efficiency, we designed the \mbox{on-chip} SPM in a way to minimize the \mbox{off-chip} memory accesses, while maintaining high throughput. 
We explored different architectural designs for our {\it DESCNet}, i.e., a specialized scratchpad memory architecture for the {\it CapsNet} accelerators. It is equipped with an application-driven memory power management unit to further reduce the leakage power. 
We performed comprehensive evaluations for both the Google's CapsNet and the DeepCaps, and illustrated significant benefits of our proposed architectures and optimizations. 
As per our knowledge, this paper proposes the first specialized memory architecture for a DNN inference accelerator executing {\it CapsNets}. 
Our work motivates the need for application-specific design and optimizations towards energy-efficient memory architectures for the next-generation embedded DNN inference hardware.

\section*{Acknowledgment}

This work has been partially supported by the Doctoral College Resilient Embedded Systems which is run jointly by TU Wien's Faculty of Informatics and FH-Technikum Wien, by the Czech Science Foundation project 19-10137S and by the the Czech Ministry of Education of Youth and Sports from the Operational Program Research, Development and Education project International Researcher Mobility of the Brno University of Technology -- CZ.02.2.69/0.0/0.0/16\_027/0008371.

\vspace*{0pt}
\begin{refsize}
\bibliographystyle{IEEEtran}
\bibliography{main.bib}
\end{refsize}

\begin{figure}[h]
	\vspace*{50pt}
\end{figure}

\begin{IEEEbiography}[{\includegraphics[width=1in]{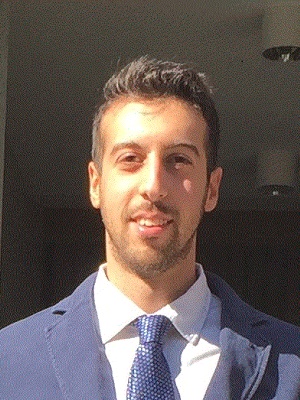}}]{Alberto Marchisio}
(S'18) received his B.Sc. degree in Electronic Engineering from Politecnico di Torino, Turin, Italy, in October 2015. He received his M.Sc. degree in Electronic Engineering (Electronic Systems) from Politecnico di Torino, Turin, Italy, in April 2018. Currently, he is Ph.D. Student at Computer Architecture and Robust Energy-Efficient Technologies (CARE-Tech.) lab, Institute of Computer Engineering, Technische Universit{\"a}t Wien (TU Wien), Vienna, Austria, under the supervision of Prof. Dr. Muhammad Shafique. He is also a student IEEE member. His main research interests include hardware and software optimizations for machine learning, brain-inspired computing, VLSI architecture design, emerging computing technologies, robust design, and approximate computing for energy efficiency. He received the honorable mention at the Italian National Finals of Maths Olympic Games in 2012, and the Richard Newton Young Fellow Award in 2019.
\end{IEEEbiography}

\begin{IEEEbiography}[{\includegraphics[width=1in]{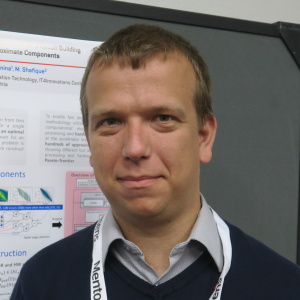}}]{Vojtech Mrazek}
(M'18) received a Ing. and Ph.D. degrees in information technology from the Faculty of Information Technology, Brno University of Technology, Czech Republic, in 2014 and 2018. He is a researcher at the Faculty of Information Technology with Evolvable Hardware Group and he is also a visiting post-doc researcher at Department of Informatics, Institute of Computer Engineering, Technische Universit{\"a}t Wien (TU Wien), Vienna, Austria. His research interests are approximate computing, genetic programming and machine learning. He has authored or co-authored over 30 conference/journal papers focused on approximate computing and evolvable hardware. He  received several awards for his research in approximate computing, including the Joseph Fourier Award in 2018 for research in computer science and engineering.

\end{IEEEbiography}

\begin{IEEEbiography}[{\includegraphics[width=1in]{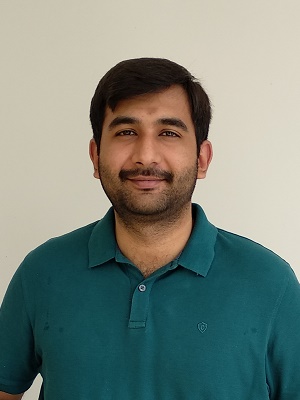}}]{Muhammad Abdullah Hanif}
received the B.Sc. degree in electronic engineering from the Ghulam Ishaq Khan Institute of Engineering Sciences and Technology (GIKI), Pakistan, and the M.Sc. degree in electrical engineering with a specialization in digital systems and signal processing from the School of Electrical Engineering and Computer Science, National University of Sciences and Technology, Islamabad, Pakistan. He is currently a University Assistant with the Department of Informatics, Institute of Computer Engineering, Technische Universit{\"a}t Wien (TU Wien), Austria.

\noindent He was also a Research Associate with the Vision Processing Lab, Information Technology University, Pakistan, and as a Lab Engineer with GIKI, Pakistan. His research interests are in brain-inspired computing, machine learning, approximate computing, computer architecture, energy-efficient design, robust computing, system-on-chip design, and emerging technologies. He was a recipient of the President’s Gold Medal for the outstanding academic performance during the M.Sc. degree.
\end{IEEEbiography}

\begin{IEEEbiography}[{\includegraphics[width=1in]{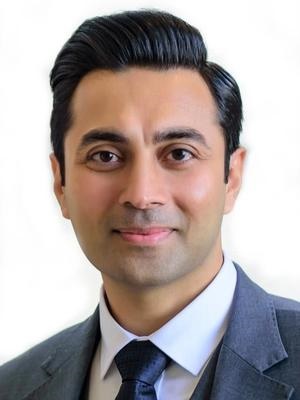}}]{Muhammad Shafique}
(M'11-SM'16) received the Ph.D. degree in computer science from the Karlsruhe Institute of Technology (KIT), Germany, in 2011. Afterwards, he established and led a highly recognized research group at KIT for several years as well as conducted impactful R\&D activities in Pakistan. In Oct.2016, he joined the Institute of Computer Engineering at the Faculty of Informatics, Technische Universit{\"a}t Wien (TU Wien), Vienna, Austria as a Full Professor of Computer Architecture and Robust, Energy-Efficient Technologies. Since Sep.2020, he is with the Division of Engineering, New York University Abu Dhabi (NYU AD), United Arab Emirates, and is a Global Network faculty at the NYU Tandon School of Engineering, USA.

\noindent
His research interests are in brain-inspired computing, AI \& machine learning hardware and system-level design, energy-efficient systems, robust computing, hardware security, emerging technologies, FPGAs, MPSoCs, and embedded systems. His research has a special focus on cross-layer analysis, modeling, design, and optimization of computing and memory systems. The researched technologies and tools are deployed in application use cases from Internet-of-Things (IoT), smart Cyber-Physical Systems (CPS), and ICT for Development (ICT4D) domains.

\noindent
Dr. Shafique has given several Keynotes, Invited Talks, and Tutorials, as well as organized many special sessions at premier venues. He has served as the PC Chair, General Chair, Track Chair, and PC member for several prestigious IEEE/ACM conferences. Dr. Shafique holds one U.S. patent has (co-)authored 6 Books, 10+ Book Chapters, and over 250 papers in premier journals and conferences. He received the 2015 ACM/SIGDA Outstanding New Faculty Award, AI 2000 Chip Technology Most Influential Scholar Award in 2020, six gold medals, and several best paper awards and nominations at prestigious conferences.
\end{IEEEbiography}

\clearpage
\appendices
\section{Off-Chip Memory Analysis}
\label{app:offchip}

\vspace*{2pt}
\noindent
\textit{1.\ \ Operation-Wise Off-Chip Analysis for the Google's CapsNet on the MNIST dataset}
\vspace*{3pt}

\begin{figure}[h]
	\centering
	\vspace*{-2pt}
	\includegraphics[width=\linewidth]{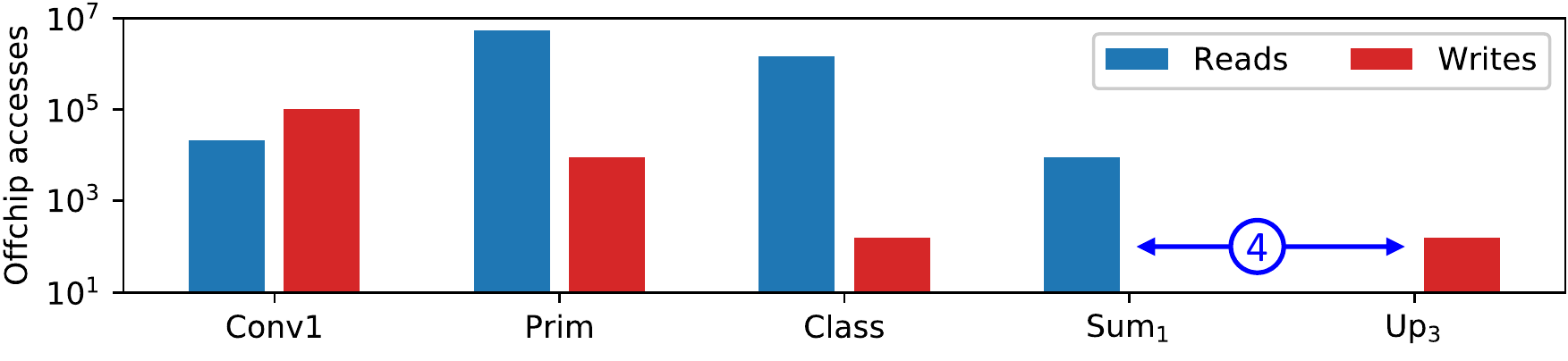}
	\vspace*{-10pt}
	\caption{Off-chip accesses for the CapsNet inference.}
	\label{fig:offc_caps}
	\vspace*{-0pt}
\end{figure}

The \mbox{off-chip} accesses\footnote{Note, before being written into the \mbox{off-chip} memory (when needed), the results coming out from the accumulators pass through the activation units. However, for ease-of-discussion, we relate these \mbox{off-chip} memory writes to the accumulators.}, reported in Fig. \ref{fig:offc_caps}, can be computed using the \Cref{eq:offchip_read,eq:offchip_write}, which are valid for the first three operations, indicated with the index $i$. 
$RD_{off}$ and $WR_{off}$ indicate the SPM read and write accesses, while the subscripts $_D$ and $_W$ stand for \mbox{on-chip} data and weight memories, respectively.
\textit{In the dynamic routing, the \mbox{off-chip} memory is not accessed, except for the first and last operation, because all the values required during the dynamic routing are stored \mbox{on-chip}.}

\vspace*{-2pt}
\begin{equation}
\vspace*{-0pt}
    \left( RD_{off} \right) _{i} = \left( WR_{D} + WR_{W} \right) _{i}
    \label{eq:offchip_read}
\end{equation}
\begin{equation}
    \left( WR_{off} \right) _{i} = \left( RD_{D} \right) _{i+1}
    \label{eq:offchip_write}
\end{equation}

Note, the peak of accesses are measured for the PrimaryCaps (Prim) layer. On the contrary, during the dynamic routing, the \mbox{off-chip} memory is not accessed, except for read accesses in the first operation and write accesses in the last one (see pointer \rpoint{4} in Fig.~\ref{fig:offc_caps}), due to the efficient data and weight reuse in these operations.

\vspace*{8pt}
\noindent
\textit{2.\ \ Operation-Wise Off-Chip Analysis for the DeepCaps on the CIFAR10 dataset}
\vspace*{3pt}

\begin{figure}[h]
	\centering
	\includegraphics[width=\linewidth]{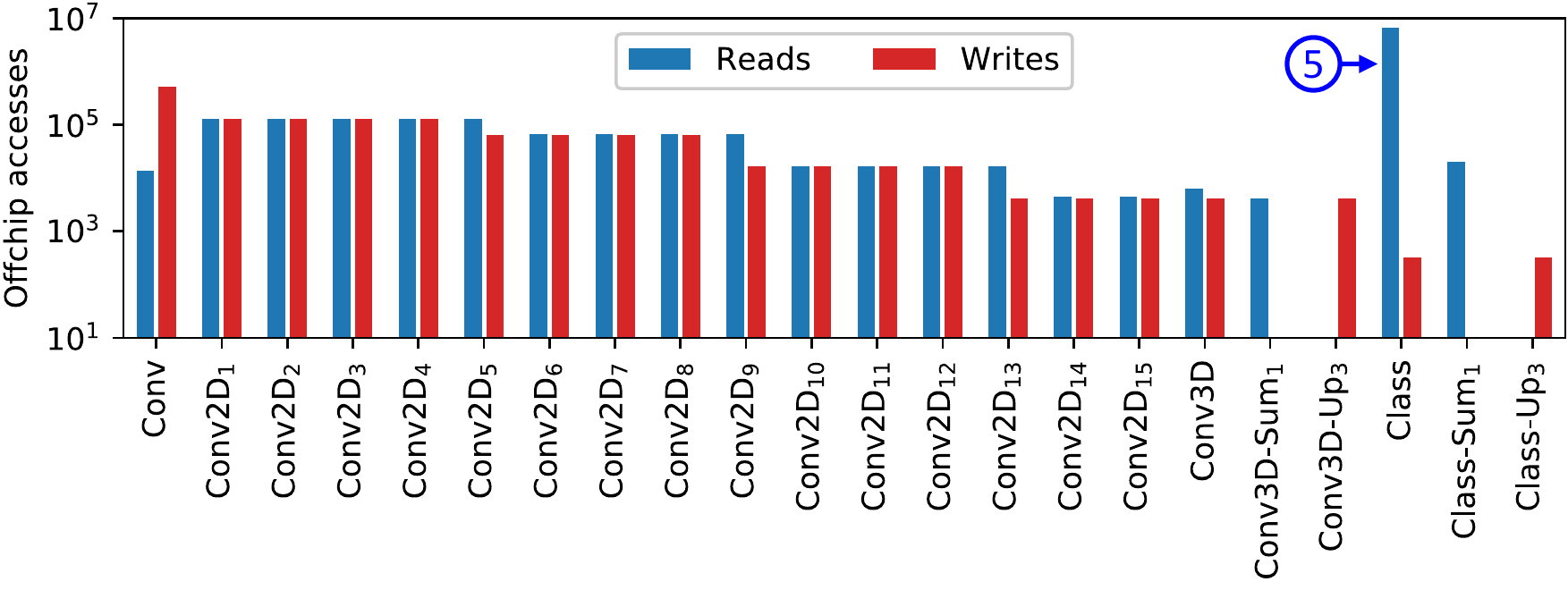}
	\vspace*{-15pt}
	\caption{Off-chip accesses for the DeepCaps inference.}
	\label{fig:offc_deepcaps}
\end{figure}

The \mbox{off-chip} accesses for the DeepCaps are shown in Fig.~\ref{fig:offc_deepcaps}. While reads and writes proportionally decrease by decreasing the sizes of the convolutional layers, for the dynamic routing the accesses are low, thanks to the efficient reuse. The peak is visible at the beginning of the \textit{ClassCaps} layer (see pointer \rpoint{5} in Fig.~\ref{fig:offc_deepcaps}), which is due to the large number of weights in that operation.

\vspace*{-8pt}
\section{Scratchpad Memory Breakdown}
\label{app:mem_breakdown}

\vspace*{3pt}
\noindent
\textit{1.\ \ Breakdown for the Selected Pareto-Optimal Solutions of the Google's CapsNet on the MNIST dataset}
\vspace*{3pt}

\begin{figure}[h]
	\centering
	\includegraphics[width=\linewidth]{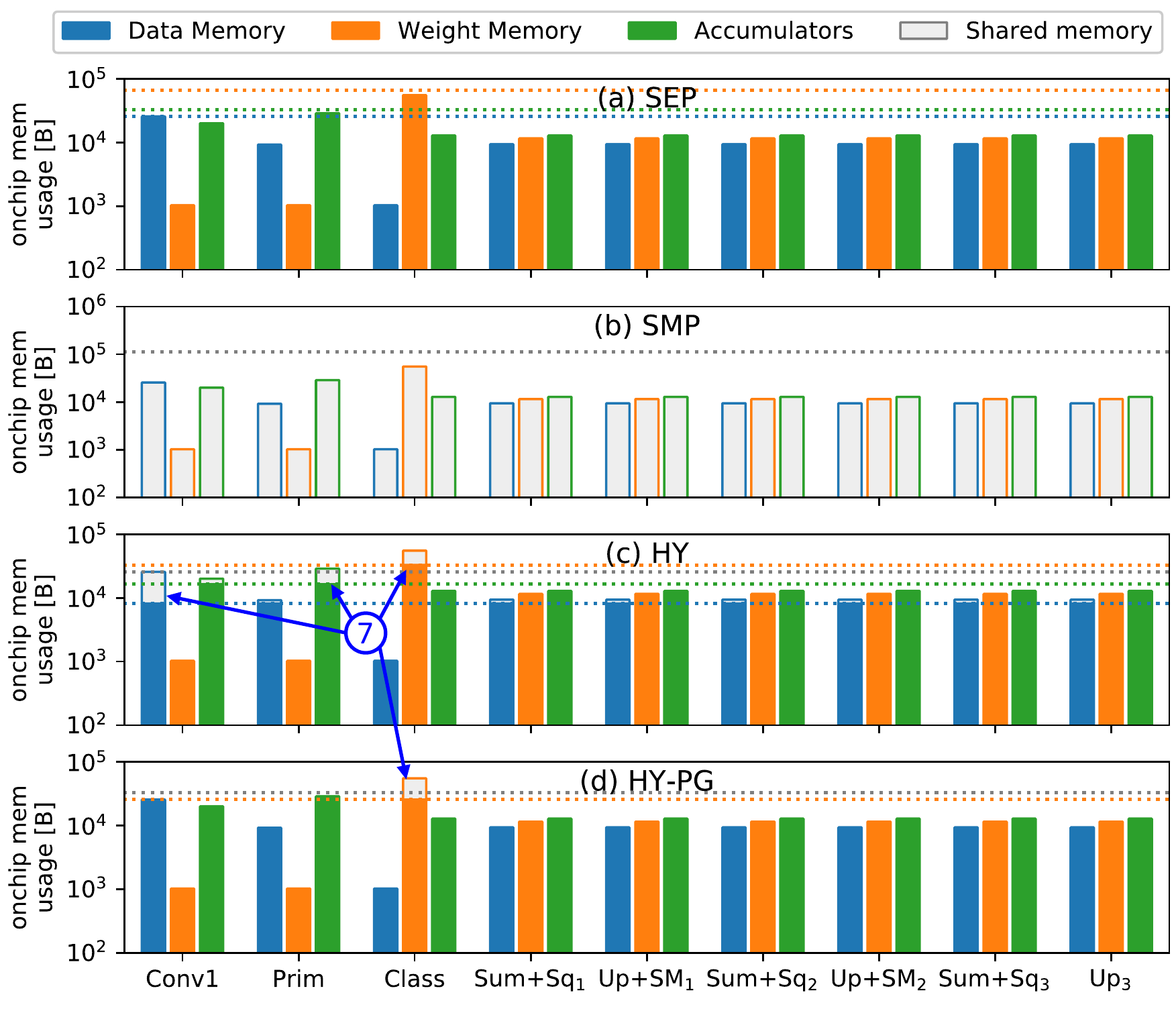}
	\caption{Memory breakdown for different design options of the \textit{DESCNet} for the CapsNet. The dashed lines show the respective memory sizes.}
	\label{fig:mems_caps}
\end{figure}

Fig.~\ref{fig:mems_caps} shows, for different design options, which type of memory is used to store the operation-wise information. 
While for the \textit{SEP} and \textit{SMP} designs the picture is clear and relatively simple, this mean of visualization is extremely useful for the hybrid solutions (\textit{HY} and\textit{ HY-PG}). It shows in a comprehensive way, for each operation, the fraction of the memory usage that is covered by the shared memory (represented with silver-colored filled bars), thereby lowering the sizes of the other memories. 
The main benefit of having an \textit{HY} configuration is that the peaks in the operation-wise memory usage can be amortized by the utilization of the shared memory (see pointer \rpoint{7} in Fig.~\ref{fig:mems_caps}).

\begin{figure}[h]
	\centering
	\includegraphics[width=\linewidth]{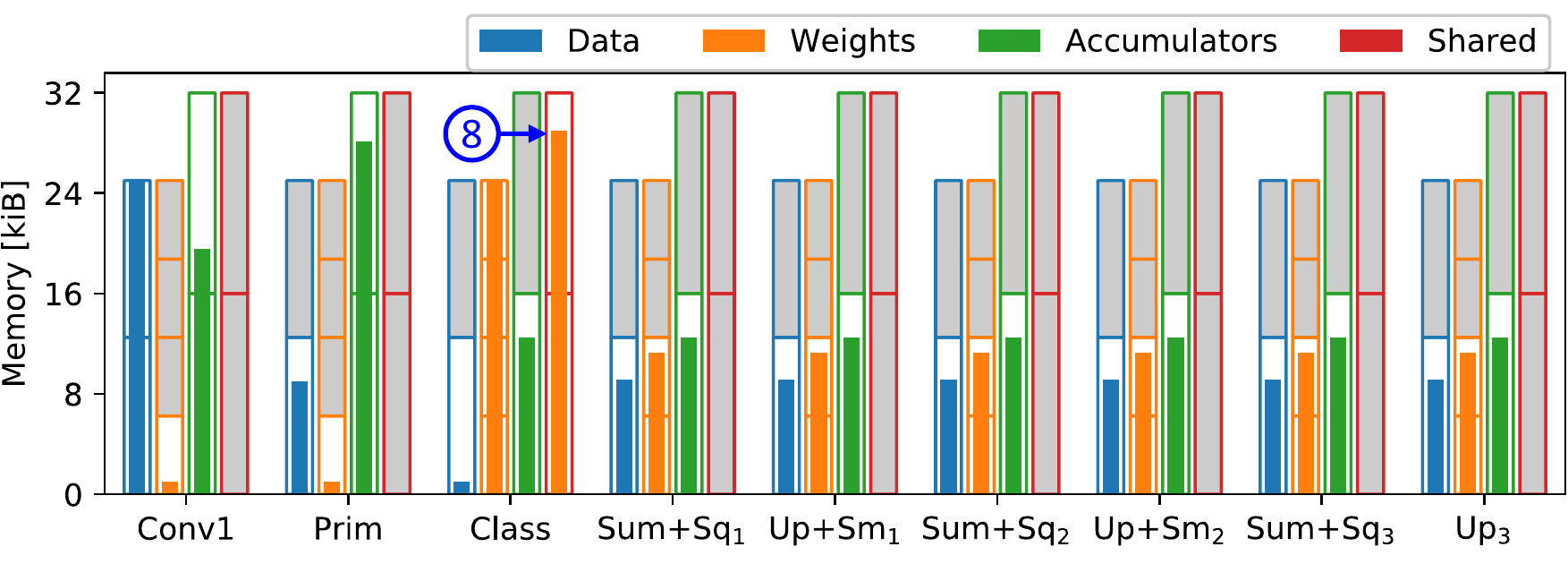}
	\caption{Illustrative example of the \mbox{power-gating}, applied to the \textit{HY-PG} design option for the Google's CapsNet.}
	\label{fig:caps_mem_banks}
\end{figure}

An example showing the \mbox{power-gating} mechanism for the \textit{HY-PG} organization is reported in Fig.~\ref{fig:caps_mem_banks}. Here, the colored boxes represent the memory sectors. The $ON$ sectors are filled in white, while the $OFF$ sectors in grey. The figure highlights that the shared memory is often powered $OFF$, except for the \textit{ClassCaps} (Class) operation, where its sectors are active and contain part of the necessary weights for this operation. (see pointer \rpoint{8} in Fig.~\ref{fig:caps_mem_banks}).

\vspace*{5pt}
\noindent
\textit{2.\ \ Breakdown for the Selected Pareto-Optimal Solutions of the DeepCaps on the CIFAR10 dataset}
\vspace*{3pt}

\begin{figure}[h]
	\centering
	\includegraphics[width=\linewidth]{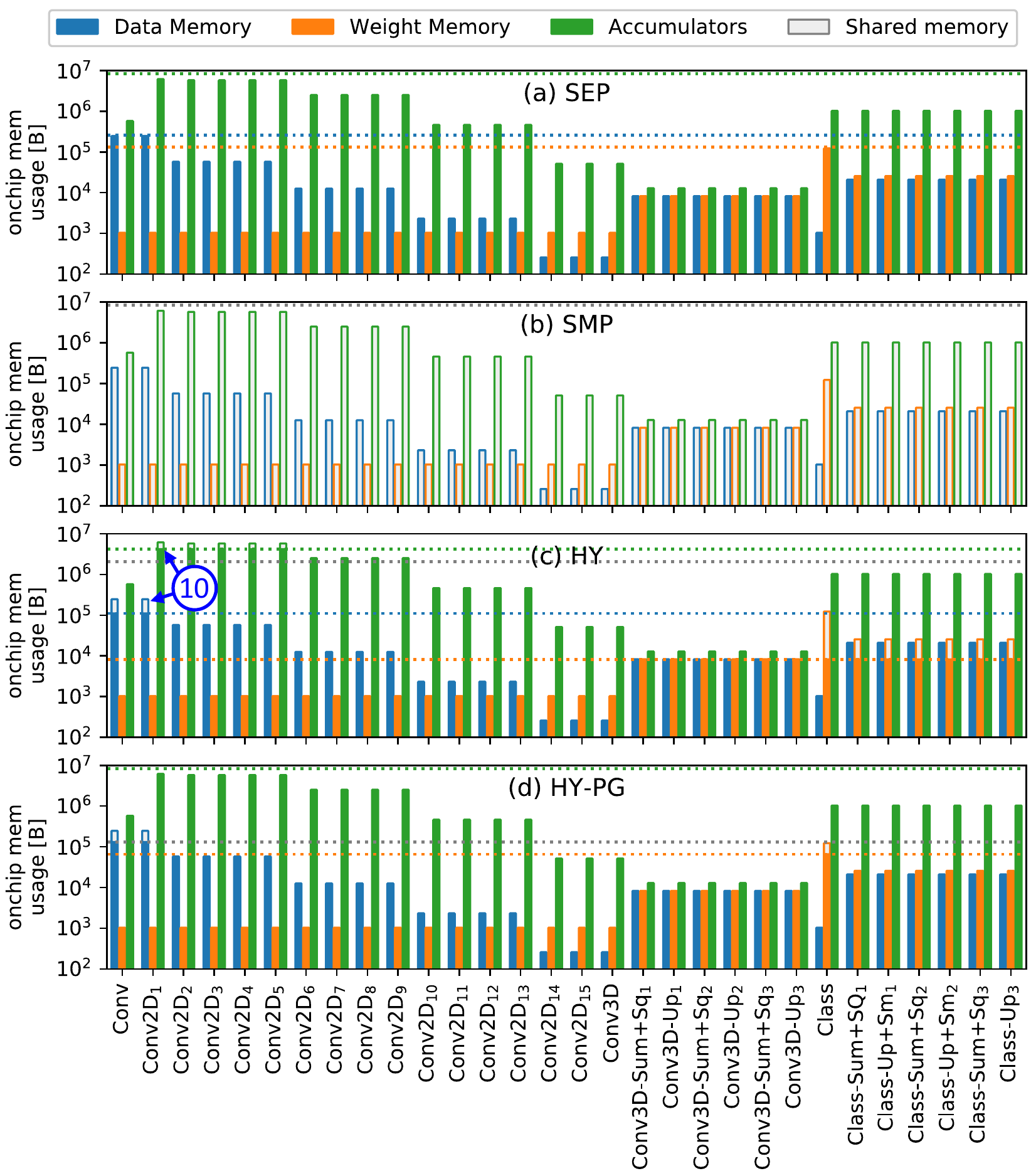}
	\caption{Memory breakdown for different design options of the \textit{DESCNet} for the DeepCaps. The dashed lines show the respective memory sizes.}
	\label{fig:mems_deepcaps}
\end{figure}

The memory breakdown showing operation-wise which type of memory is used is shown in Fig.~\ref{fig:mems_deepcaps}. Note, the analysis shows that the shared memory of the \textit{HY} and \textit{HY-PG} design options do not always require having three ports, because for some solutions, the shared memory only needs to store one or two different types of values. For example, as shown by pointer \rpoint{10} in Fig.~\ref{fig:mems_deepcaps}c, the shared memory covers simultaneously part of the data and part of the accumulator memory usage for the operation $CONV2D_1$, while, for all the other operations of the DeepCaps, at most one type of value is covered by the shared memory. Therefore, for this particular solution, a 2-port shared memory is sufficient to guarantee the correct functioning of the system, thus potentially saving area and energy, compared of having an equivalent 3-port memory.

\begin{figure}[t]
	\centering
	\includegraphics[width=\linewidth]{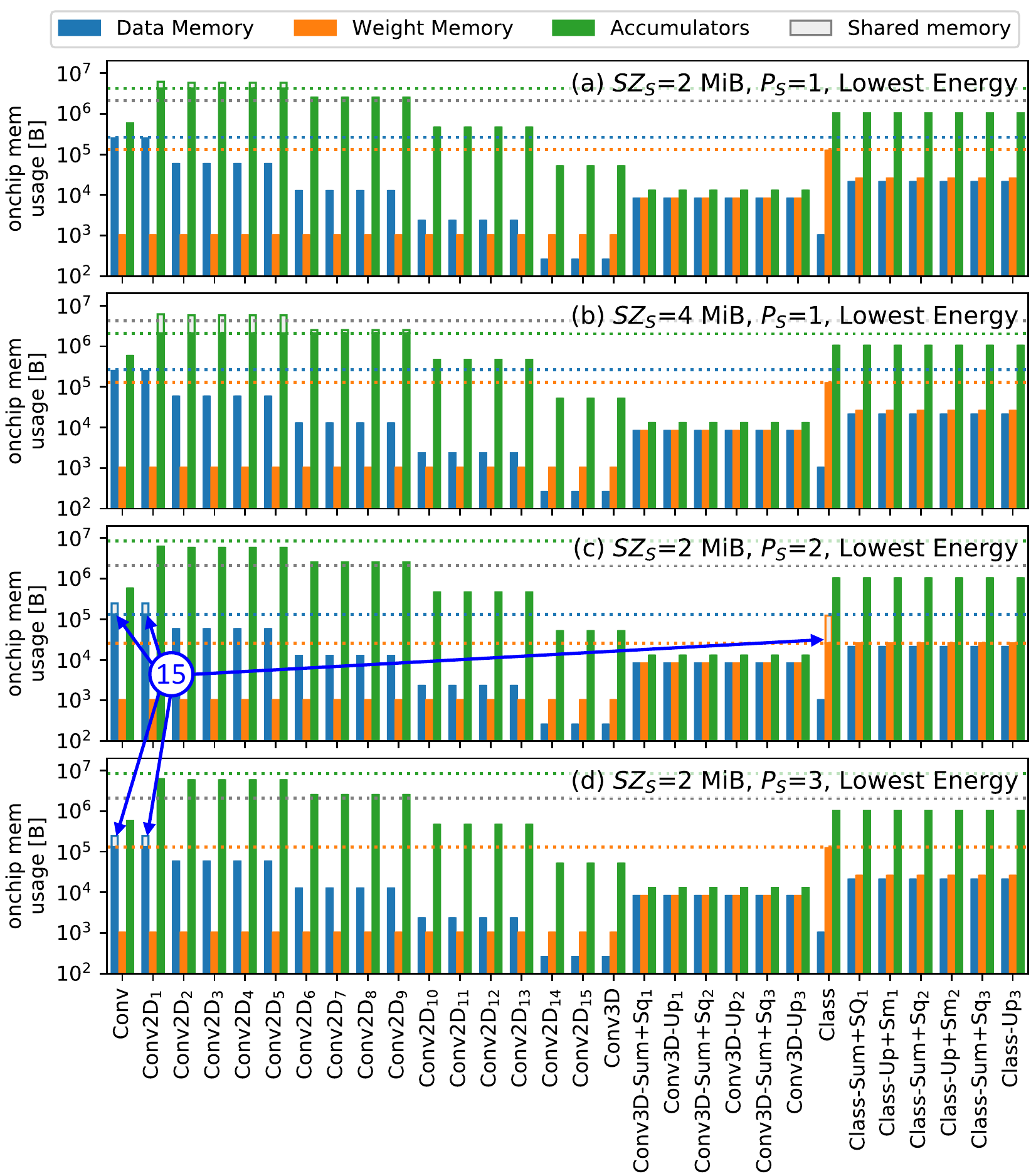}
	\caption{Memory breakdown for the \textit{HY-PG DESCNet} design option with different constraints on the shared memory for the DeepCaps. The dashed lines show the respective memory sizes.}
	\label{fig:mems_deepcaps_dc}
	\vspace*{350pt}
\end{figure}

Therefore, this outcome motivated us to conduct the exploration of the space of the \textit{HY-PG} solutions with constraints on the number of ports of the shared memory, as discussed in Section~\ref{subsec:DSE_constraint}. Fig.~\ref{fig:mems_deepcaps_dc} shows the memory breakdown for the lowest-energy \textit{HY-PG} solutions with the given constraints on the shared memory. The 1-port shared memory of sizes $2~MiB$ and $4~MiB$ can partially contain accumulator values for some DeepCaps' operations, while the lowest-energy solutions for the 2-port and 3-port shared memory of size $2~MiB$ partially contain the weights and data values (see pointer \rpoint{15} in Fig.~\ref{fig:mems_deepcaps_dc}). Hence, for the latter two solutions, the size of the accumulator memory is not reduced.

\end{document}